\documentclass[sigconf,nonacm]{acmart}
\AtBeginDocument{%
  }

\usepackage{pifont}

\NewDocumentCommand{\var}{O{s} m O{}}{%
  \ensuremath{#1_{#2}^{#3}}
}
\usepackage{siunitx}
\usepackage{colortbl}
\usepackage{algorithm}

\usepackage{amsmath}
\usepackage{listings}
\usepackage{xcolor}  
\usepackage{subfig}
\usepackage{footmisc}  

\usepackage{microtype} 

\usepackage{algpseudocode}
\usepackage{xspace}

\definecolor{codered}{rgb}{0.6,0,0}
\definecolor{codegreen}{rgb}{0,0.6,0}
\definecolor{codeblue}{rgb}{0,0,0.6}
\definecolor{codepurple}{rgb}{0.58,0,0.82}
\definecolor{codebg}{rgb}{0.97,0.97,0.97}

\newcommand{\commentout}[1]{}

\definecolor{light-gray}{gray}{0.80}

\newcommand\fref{Fig.~\ref}

\newcommand\sref{\S~\ref}

\usepackage{amsthm}

\usepackage{amsthm}

\newtheorem{mytheorem}{Theorem}
\newtheorem{assumption}[mytheorem]{Assumption}
\newtheorem{lemma}[mytheorem]{Lemma}
\newtheorem{remk}[mytheorem]{Remark}

\newcommand{\name}{X-MoE\xspace}
\newcommand{\mconv}{$M_{conv}$\xspace}
\newcommand{\mspec}{$M_{spec}$\xspace}

\newcommand{\rbd}{RBD\xspace}
\newcommand{\ssmb}{SSMB\xspace}

\newcommand{\metadata}{ERI-arrays\xspace}
\newcommand{\metadatum}{ERI-array\xspace}

\newcommand{\adispatch}{$A_{\texttt{dispatch}}$\xspace}
\newcommand{\acombine}{$A_{\texttt{combine}}$\xspace}

\newcommand{\ainterf}{$A^0_{\texttt{interm}}$\xspace}
\newcommand{\ainters}{$A^1_{\texttt{interm}}$\xspace}
\usepackage[skip=2pt]{caption}
\usepackage{setspace}
\usepackage{enumitem}
\usepackage{titlesec}
\usepackage{needspace}
\usepackage{etoolbox}

\makeatletter
\patchcmd{\@afterheading}{\@nobreaktrue}{\@nobreakfalse}{}{}
\makeatother

\makeatletter 
\def\thm@space@setup{%
  \thm@preskip=3pt
  \thm@postskip=\thm@preskip 
}
\makeatother

\setlist[itemize]{noitemsep, topsep=0pt}

\makeatletter
\g@addto@macro\normalsize{%
  \setlength\abovedisplayskip{1pt}
  \setlength\belowdisplayskip{1pt}
  \setlength\abovedisplayshortskip{1pt}
  \setlength\belowdisplayshortskip{1pt}
}
\makeatother

\begin{document}

\title{\name: Enabling Scalable Training for Emerging Mixture-of-Experts Architectures on HPC Platforms}


\author{Yueming Yuan}
\affiliation{
  \institution{UIUC}
  \city{Urbana}
  \state{IL}
  \country{USA}}
\email{yy28@illinois.edu}

\author{Ahan Gupta}
\affiliation{
  \institution{UIUC}
  \city{Urbana}
  \state{IL}
  \country{USA}}
\email{ag82@illinois.edu}

\author{Jianping Li}
\affiliation{
  \institution{UIUC}
  \city{Urbana}
  \state{IL}
  \country{USA}}
\email{jli199@illinois.edu}

\author{Sajal Dash}
\affiliation{
  \institution{Oak Ridge National Laboratory}
  \city{Oak Ridge}
  \state{TN}
  \country{USA}}
\email{dashs@ornl.gov}

\author{Feiyi Wang}
\affiliation{
  \institution{Oak Ridge National Laboratory}
  \city{Oak Ridge}
  \state{TN}
  \country{USA}}
\email{fwang2@ornl.gov}

\author{Minjia Zhang}
\affiliation{
  \institution{UIUC}
  \city{Urbana}
  \state{IL}
  \country{USA}}
\email{minjiaz@illinois.edu}
\renewcommand{\shortauthors}{Trovato et al.}

\begin{abstract}

Emerging expert-specialized Mixture-of-Experts (MoE) architectures, such as DeepSeek-MoE, deliver strong model quality through fine-grained expert segmentation and large top-k routing. However, their scalability is limited by substantial activation memory overhead and costly all-to-all communication. Furthermore, current MoE training systems -- primarily optimized for NVIDIA GPUs -- perform suboptimally on non-NVIDIA platforms, leaving significant computational potential untapped.
In this work, we present X-MoE, a novel MoE training system designed to deliver scalable training performance for next-generation MoE architectures. X-MoE achieves this via several novel techniques, including efficient padding-free MoE training with cross-platform kernels, redundancy-bypassing dispatch, and hybrid parallelism with sequence-sharded MoE blocks. Our evaluation on the Frontier supercomputer, powered by AMD MI250X GPUs, shows that X-MoE scales DeepSeek-style MoEs up to 545 billion parameters across 1024 GPUs -- 10x larger than the largest trainable model with existing methods under the same hardware budget, while maintaining high training throughput. The source code of X-MoE is available at \url{https://github.com/Supercomputing-System-AI-Lab/X-MoE}.

\end{abstract}

\maketitle

\section{Introduction}
\label{sec:intro}

Large Language Models (LLMs) have become the backbone of modern AI applications, achieving remarkable results across domains such as dialogue systems, code generation, and scientific reasoning~\cite{gpt-3,gpt-4,openai2024gpt4ocard}. However, training these models at scale remains prohibitively expensive. For example, training models at the scale of GPT-3 or GPT-4 consumes hundreds of thousands of GPU days and incurs billions of dollars in compute cost~\cite{llama3,google-gemini,claude,xai}. As such, reducing the training cost while still achieving high model quality has become a key research challenge. 

Numerous efforts have been made to improve the training efficiency of LLMs. 
Among those, Mixture-of-Experts (MoE) have emerged as a promising path to enable sublinear compute 
with respect to the model parameters, allowing for improved model quality without increased training costs~\cite{shazeer2017outrageously,lepikhin2020gshard,fedus2022switch,jiang2024mixtral,llama4}. 
In contrast to dense models, MoEs sparsely activate model parameters, which allows one to scale to larger model parameters while keeping per-token compute budgets relatively low. 
Prior works have shown that MoEs can successfully scale to trillions of parameters~\cite{fedus2022switch}.

More recently, models such as DeepSeek-MoE~\cite{deepseek-moe} represent a new class of emerging MoE architectures that depart from earlier designs like GShard~\cite{g-shard} and Mixtral-MoE~\cite{jiang2024mixtral}. 
These models rely on architectural modifications such as fine-grained experts and large top-$k$ routing to allow experts to focus on more distinct contextual concepts, known as expert specialization. 
As a result, DeepSeek-MoE style models have great potential to accelerate LLM training with low costs, renewing interest in developing scalable and efficient training systems for emerging MoE architectures.

Unfortunately, training expert-specialized MoEs at scale is very challenging. First, existing MoE training systems heavily rely on CUDA-specific implementations for standard MoEs, which are inefficient for expert-specialized MoEs and difficult to port to non-NVIDIA platforms such as AMD Instinct GPUs or Slingshot-based interconnects using ROCm and RCCL. This lack of cross-platform support leads to inflated memory usage and suboptimal performance on heterogeneous HPC systems like Frontier~\cite{frontier} and Aurora~\cite{aurora} (\sref{subsec:lack-cross-platform}). Second, expert-specialized MoEs introduce a structural shift: they increase the number of routed experts per token and shrink each expert's hidden dimension. This change shifts the memory bottleneck from model parameters to activations, particularly in the dispatch and combine stages. However, existing MoE training systems, such as DeepSpeed-MoE~\cite{deepspeed-moe}, DeepSpeed-TED~\cite{deepspeed-ted}, and Tutel~\cite{hwang2023tutel}, do not effectively address this shifted bottleneck, causing a memory explosion (\sref{subsec:bottleneck-shift}).
Third, many-expert routing significantly increases duplication in communication, especially when multiple experts are selected per token. On platforms with hierarchical interconnects, such as Dragonfly network~\cite{dragonfly} in Frontier, this results in inefficient use of inter-node bandwidth and causes communication to become a major training efficiency bottleneck as the expert granularity increases (\sref{subsec:communication_analysis}).

To address these challenges, we present system-level optimizations for training emerging MoE architectures on cross-platform, non-NVIDIA hardware. Our analysis reveals that off-the-shelf MoE training systems, designed under the assumption of conventional MoEs and NVIDIA platforms, perform sub-optimally on new MoE architectures and non-NVIDIA hardware. For example, we observe that state-of-the-art MoE frameworks like Tutel~\cite{hwang2023tutel} and DeepSpeed-MoE~\cite{deepspeed-moe} achieve $<10$ TFLOPS on AMD MI250X GPUs, which is under 10\% of their peak performance, whereas Megablocks\allowbreak~\cite{gale2023megablocks} is highly integrated with NVIDIA Megatron-LM, which does not easily run on AMD hardware. 
Motivated by these observations, we propose \name, an MoE training system that enables scaling of expert-specialized MoEs across massive GPUs. 
\name accomplishes this via a combination of techniques,
such as padding-free MoE training with cross-platform kernels for improved memory and communication efficiency(\sref{subsec:kernels}), redundancy-bypassing dispatching for communication reduction (\sref{subsec:rbd}), and hybrid parallelism with sequence-sharded MoE blocks (\sref{subsec:tsp}). 
Unlike prior systems, we achieve this without relying on vendor-specific software stacks like CUDA, and instead rely on portable backends such as Triton~\cite{triton}. This makes \name backend-agnostic and suitable for future hardware platforms. 
To our knowledge, \name is the first work that systematically optimizes for both emerging expert-specialized MoEs and the heterogeneity of non-NVIDIA platforms.

We demonstrate our approach on the Frontier supercomputer~\cite{frontier}, which consists of AMD MI250X GPUS with Dragonfly architecture~\cite{dragonfly}. We validate our design through comprehensive experiments. 
\name enables the training of DeepSeek-style MoEs up-to 545B-parameters on 1024 AMD GPUs, which is 10$\times$ larger than the largest trainable model under the same hardware budget using existing solutions. 
Beyond model scale, \name delivers up to 1.42x higher training throughput than state-of-the-art MoE systems, while also outperforming them in both weak and strong scaling.  
To enhance usability, \name has been integrated with DeepSpeed~\cite{deepspeed}, a popular open-source DL training library, making it accessible for future MoE training workloads.

\section{Background}
\label{sec:background}


In this part, we introduce important concepts in the MoE literature.

\begin{figure}[t]
    \centering
    \includegraphics[width=1\linewidth]{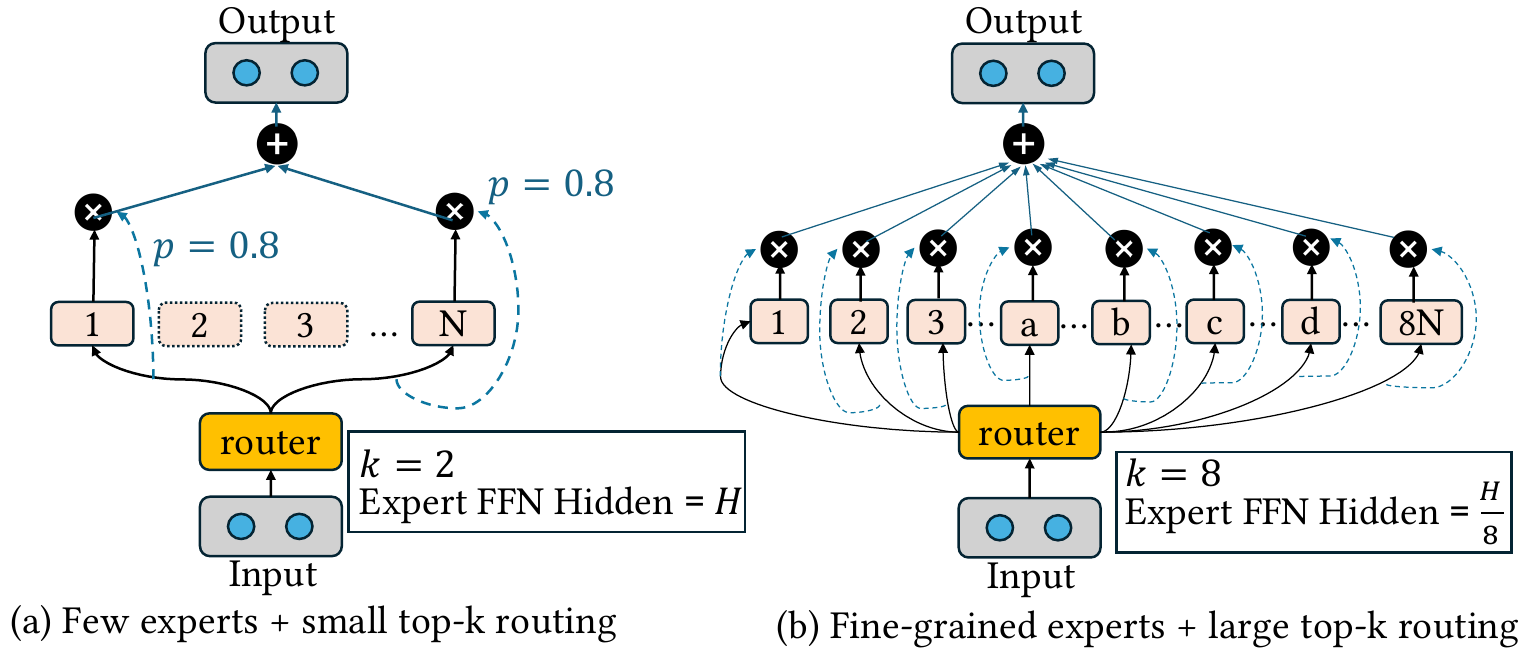}
    \caption{Standard vs. expert-specialized MoE architecture.}    
     \Description{ }
    \label{fig:moe-comparison}
\end{figure}

\noindent
\textbf{\emph{Sparsely Activated MoEs.}}
\label{subsec:moe}
An MoE layer most commonly replaces the dense FFN in a transformer with a single layer consisting of a set of experts~\cite{shazeer2017outrageously,fedus2022switch}. During training, a token is dynamically routed to \textit{k} experts based on scores computed by a gating function. This is done in four steps. First, the gating function is applied to the input tokens, generating a mapping of which token should be processed by which expert. Second, a dispatching stage is responsible for routing each token to its respective mapped experts' input buffers. Note, experts could reside on different devices, requiring the need for an all-to-all to exchange tokens between devices. Third, each expert independently processes all the tokens mapped to it. Fourth, a combine stage re-routes the tokens back to their original device via a second all-to-all, combining all the expert outputs and reordering the tokens to match the order of the original tokens input to the layer. \fref{fig:moe-comparison}(a) illustrates the MoE architecture.

\noindent
\textbf{\emph{Expert-Specialized Mixture-of-Experts.}}
\label{subsec:deepseek}
The MoE architecture design paradigm has recently evolved. Prior to DeepSeek-MoE~\cite{deepseek-moe}, state-of-the-art MoEs like Mixtral-MoE use a small number of experts (e.g., 8), each of which is similar to the FFN layers in corresponding dense models with a small top-k gating value (e.g., $k$ is typically $1$ or $2$). However, such models are shown to lack expert specialization~\cite{jiang2024mixtral}. To overcome the issue, DeepSeek-MoE introduces both fine-grained experts and large top-k routing to increase expert specialization (\fref{fig:moe-comparison}(b)). Instead of a few large experts, the model uses a much higher number of smaller experts and activates more of them for reach token. Concretely, each expert's hidden dimension is reduced to a fraction (e.g., $1/m$, where $m\propto k$) of the size used in a standard MoE's FFN. Meanwhile, the number of experts increases roughly $m$-fold. 
For example, if a standard MoE has an FFN dimension of 4096 with 8 experts (activating 2 per token), DeepSeek-MoE splits this into $m=8$ fine-grained experts per original expert dimension, yielding 64 experts total and activating 16 per token. This keeps the total parameters and per-token computation roughly the same as before but dramatically expands the combination of experts a token can see from only 28 possible expert-pair combinations in the original example to nearly $4.89\times10^{14}$ combinations. DeepSeek-MoE models employ on the order of hundreds of experts per MoE layer, e.g., DeepSeek-v3 uses 256 experts in each layer, with 8 experts activated for every token, which exhibits far more expressive power than standard MoEs with the same parameter budget.

\noindent
\textbf{\emph{Existing MoE Training Frameworks.}}
\label{subsec:moe-frameworks}
Training MoE models at scale introduces a number of system-level challenges. Early systems, such as GShard, introduce Expert Parallelism (EP)~\cite{g-shard}, which enables efficient MoE scaling across devices by distributing experts across GPUs. Subsequently, several large-scale MoE training frameworks were introduced~\cite{deepspeed-moe,deepspeed-ted,gale2023megablocks,faster-moe,lancet,brainstorm}. DeepSpeed-MoE combines ZeRO-style Data Parallelism (ZeRO-DP) with EP, offering more memory-efficient training for large-scale MoE models. A more recent extension of DeepSpeed-MoE, DeepSpeed-TED~\cite{deepspeed-ted}, introduces three-dimensional parallelism by combining DP, EP, and Tensor-slicing Parallelism (TP) to scale MoEs further. However, they are designed for low top-k values and coarse-grained experts. Tutel~\cite{hwang2023tutel} proposes an adaptive DP and TP strategy, optimizing memory and compute usage by dynamically switching between data and tensor parallelism depending on the load distributed amongst experts. 
Recently, Megablocks~\cite{gale2023megablocks} introduces sparse primitives and a no-token dropping scheme to process MoE layers, representing everything as block-sparse matrix multiplications. However, in doing so, their kernel requires padding the token buffer to multiples of a preset size, incurring serious zero-paddings on the emerging MoE workload. 

\section{Challenges and Opportunities}
\label{sec:challenges}

Expert-specialized MoEs represent a significant advancement in LLMs. However, training these emerging MoE architectures poses significant challenges for existing off-the-shelf MoE training solutions, especially on HPC platforms.
\subsection{Lack of Efficient Cross-Platform Kernels for Scaling Expert-Specialized MoEs}
\label{subsec:lack-cross-platform}
Existing MoE training frameworks often rely on a dense and static tensor layout for MoE gating and dispatching, which becomes inefficient when applied to expert-specialized MoEs. 
For example, MoE training frameworks (e.g., GShard~\cite{g-shard}, Fairseq~\cite{fairseq} and DeepSpeed-MoE~\cite{deepspeed-moe}) often implement each stage of the MoE pipeline via fast batched matrix multiplication (matmul) primitives. However, these primitives place a constraint: requiring the same number of tokens routed to each expert, which does not hold during training. To handle the dynamic assignment of tokens to experts, these frameworks introduce a fixed expert capacity $C$ and pad unused slots with zeros when fewer than $C$ tokens are routed to an expert. Conversely, tokens are dropped when capacity is exceeded. During the gating stage, a dispatching mask, \texttt{dispatch\_mask} of size \texttt{[S, E, C]} is constructed. The entry \texttt{dispatch\_mask[t, e, c]} is either 1 or 0 indicating if the \texttt{t$^{\text{th}}$} token is routed to the \texttt{c$^{\text{th}}$} position in expert \texttt{e}'s buffer. A token-dropping mask is applied over \texttt{dispatch\_mask} to additionally drop tokens. 

\begin{figure}[!ht]
    \centering
    \includegraphics[width=0.9\linewidth]{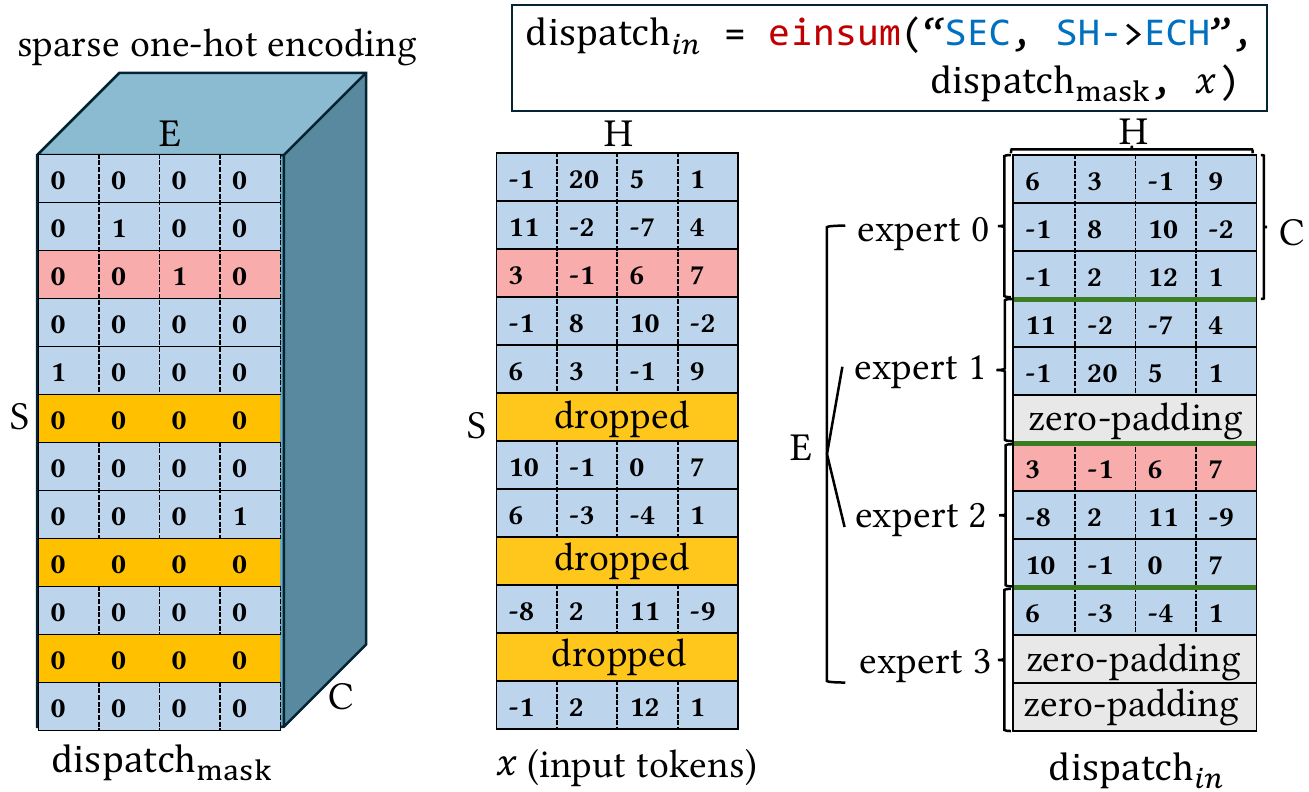}    
    \caption{Conventional gating and dispatching logic with zero-padded and large intermediate memory cost. 
    }
    \label{fig:gshard}
    \Description{}
\end{figure}

Each of the dispatch, MLP and combine stages leverage matmuls on input token-buffers to process tokens. During the dispatch stage, each worker uses an \texttt{einsum} operation on the dispatching mask and input tokens to correctly place each token into its respective experts buffers. The expert buffers are \texttt{[E, C, H]}-sized. If less than $C$ tokens are placed in an expert's buffer, the unused slots are zero-padded. \fref{fig:gshard} illustrates the dispatch process and how zero-padding is introduced into an expert's input-buffer at this stage. Next, an even \texttt{alltoall} communication exchanges token buffers across devices, correctly routing each token to the devices its experts reside on. Each expert then operates on its input token-buffers in parallel. Finally, another \texttt{alltoall} re-exchanges the tokens to their original device to generate the final output of the layer. Importantly, the zero-padding introduced in the dispatch stage is retained across each \texttt{alltoall} and expert compute. This increases both the communication volume and activation memory of MoE training.

In expert-specialized MoEs, where hundreds of fine-grained experts are activated per MoE layer and top-$k$ routing is large, the above zero-padded pipeline becomes a major memory bottleneck. In our experiments with DeepSeek-style MoEs, we observe that the dispatch mask and intermediate buffers consume over 70\% of the total activation memory when training with DeepSpeed-MoE~\cite{deepspeed-moe} using $1.25\times$ average perceived tokens per-expert as the expert capacity. This not only increases the pressure on GPU memory but also the communication volume of \texttt{alltoall} calls dependent on these buffers. While it is possible to build sparse CUDA kernels to address this issue~\cite{hwang2023tutel}, CUDA kernels are tightly coupled to NVIDIA's CUDA backend and cannot be easily ported to other platforms. The lack of cross-platform support becomes a critical limitation on different HPC systems. As a result, the current training pipeline either falls back to inefficient PyTorch-based implementations or requires costly kernel re-engineering (e.g., ROCm-based kernels) for the particular hardware target in question, which is error-prone. As MoE architectures grow more complex, the need for portable, hardware-agnostic MoE kernels becomes very essential.

\vspace{3pt}
\noindent \fbox{\parbox{0.96\linewidth}{
\textbf{Takeaway-1:} Existing MoEs rely on dense, CUDA-specific implementations that are inefficient for expert-specialized MoEs and difficult to port to non-NVIDIA platforms. This lack of cross-platform support leads to inflated memory usage and degraded performance of MoE training on heterogeneous HPC platforms.}}
\vspace{1pt}

\vspace{-1mm}
\subsection{Memory Bottleneck Shift in Expert-Specialized MoEs}
\label{subsec:bottleneck-shift}

We analyze the unique memory behavior of emerging expert-special\-ized MoE architectures by comparing them with size-equivalent conventional MoEs, where we keep the total parameters and per-token activated parameters the same. We refer to these as $M_{conv}$ (conventional MoE) and $M_{spec}$ (expert-specialized MoE). 
Table~\ref{tab:conventional-vs-emerging} summarizes the key differences. We use $E$ to denote the number of experts, $H$ for the model dimension, and $H_{FFN}$ for the intermediate hidden dimension in the expert FFN layers. We use $m$ to denote \emph{fine-grained factor}, which indicates how many fine-grained experts in $M_{spec}$ together replace a single expert in $M_{conv}$. For example, in most conventional MoEs~\cite{fedus2022switch,deepspeed-moe,jiang2024mixtral}, m = 1; in DeepSeek~\cite{deepseek-v3}, m=8 because expert FFNs have a $\sim$8$\times$ smaller hidden dimension and each token activates k=8 experts.

\begin{table}[H]
\centering
\begin{tabular}{l|cccc|cc}
\toprule
Model & \textbf{$E$} & \textbf{$H$} & \textbf{$H_{FFN}$} & $k$ & Param & Activated Param \\
\midrule
base  & - & $h$ & $h'$         & - & $2h'h$ & $2h'h$ \\
$M_{conv}$               & $e$ & $h$ & $h'$         & 1 & $2eh'h$ & $2h'h$ \\
$M_{spec}$               & $e\cdot m$ & $h$& $h'/m$ & m & $2eh'h$ & $2h'h$ \\
\bottomrule
\end{tabular}
\caption{The model configurations of size-equivalent conventional MoE $M_{conv}$ and expert-specialized MoE $M_{spec}$.}
\label{tab:conventional-vs-emerging}
\end{table}

\noindent
Training memory consists of model states and activations. Since \mconv and \mspec are size-equivalent, their model state size is identical. However, their activations behave very differently. Each MoE layer often instantiates four key activations during training:
\begin{itemize}
    \item \adispatch: dispatched input to experts;
    \item \ainterf and \ainters: intermediate activations between expert FFN sub-layers;
    \item \acombine: expert outputs before combining.
\end{itemize}

\noindent
All four tensors scale with batch size $b$, sequence length $s$, and routing factor $k$. Among them, \adispatch and \acombine scale with the model dimension $H$, while \ainterf and \ainters scale with the expert hidden dimension $H_{FFN}$. However, given that $H_{FFN}$ in \ainterf and \ainters shrink by 1/m in \mspec and $k\propto m$, \ainterf and \ainters remain constant across \mconv and \mspec. Instead, \adispatch and \acombine grow linearly with $m$. Therefore, in expert-specialized MoEs, activation memory is primarily dominated by the dispatch and combine stages. Table~\ref{tab:activation-size} summarizes this. This finding is interesting because prior work often assume that the intermediate FFN activation is large (e.g., 4H) for \mconv. Our analysis reveals that this assumption no longer holds true for expert-specialized MoEs. 

\begin{table}[!ht]
\centering
\begin{tabular}{l|cccc}
\toprule
Activation & \adispatch & \acombine & \ainterf & \ainters \\
\midrule
Tensor Size               & $kBSH$ & $kBSH$ & $kBSH_{FFN}$  & $kBSH_{FFN}$ \\
\midrule
\mconv               & $bsh$ & $bsh$ & $bsh'$  & $bsh'$ \\
\mspec               & $\textcolor{red}{m}bsh$ & $\textcolor{red}{m}bsh$& $bsh'$ & $bsh'$  \\
\bottomrule
\end{tabular}
\caption{The activation size of equivalent conventional MoE model \mconv and expert-specialized MoE model \mspec.}
\label{tab:activation-size}
\end{table}

\begin{figure}[!ht]
    \centering
    \includegraphics[width=1\linewidth]{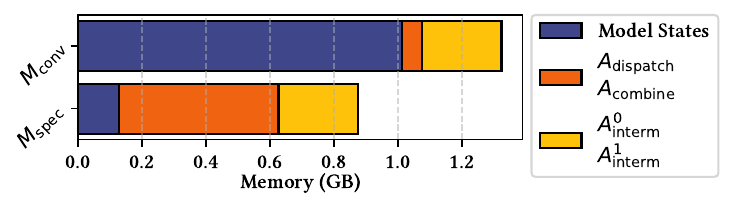}
    \caption{The MoE layer memory distribution of \mconv and \mspec created by a 6.7B base model~\cite{gpt-3} with $e=16$ and $m=8$.}
    \label{fig:conv_vs_spec_memory}
    \Description{}
\end{figure}

We further validate this observation in \fref{fig:conv_vs_spec_memory}, which compares the per-GPU memory consumption of one \mconv and \mspec layer when trained with ZeRO-1 DP and EP on 256 GPUs, using an EP size (i.e., expert-parallel group size) equal to the number of experts. The results show a clear shift in memory bottleneck. 

\noindent \fbox{\parbox{0.96\linewidth}{
\textbf{Takeaway-2:} As expert-specialized MoEs increase the number of routed experts and shrink expert hidden dimension, per-device memory bottlenecks shift from model parameters to activations, particularly in the dispatch and combine stages.
}}

\subsection{Expensive All-to-All on HPC Platform with Heterogeneous and Hierarchical Network}
\label{subsec:communication_analysis}

Most prior MoE training systems were designed for clusters composed of NVIDIA DGX nodes~\cite{deepspeed-moe,hwang2023tutel}, where nodes are connected via high-bandwidth, low-latency InfiniBand. 
These clusters exhibit relatively balanced GPU-to-GPU inter/intra-node communication bandwidths, with intra-node bandwidths only 3$\times$ faster than inter-node bandwidths. 
As a result, existing MoE systems often take advantage of this balanced network and treat all GPUs in a cluster equivalently. For example, in DeepSpeed-MoE~\cite{deepspeed-moe}, each collective involves all GPUs within an expert-parallel group, regardless of physical placement. 

However, many HPC platforms differ from DGX-style clusters. For example, Frontier~\cite{frontier} adopts a Dragonfly topology, where GPUs within a node are connected via Infinity Fabric (up to 200 GB/s) while inter-node communication happens via Slingshot (25 GB/s). This introduces a significant bandwidth asymmetry. In such hierarchical interconnects, network-aware communication is essential. Unfortunately, existing MoE systems do not exploit this hierarchy and instead route tokens indiscriminately, often leading to sub-optimal bandwidth utilization. 

\begin{figure}[!ht]
    \centering
    \includegraphics[width=\linewidth]{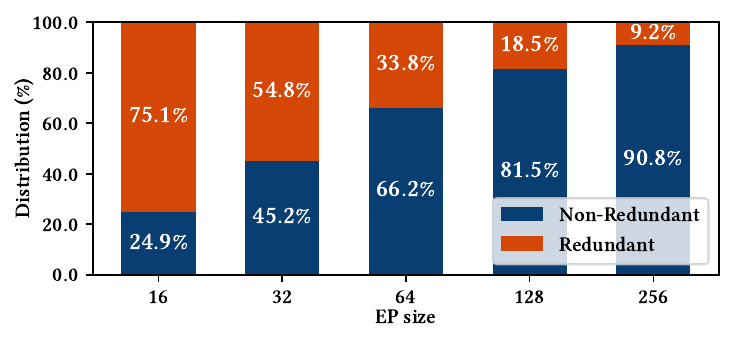}
    \caption{Redundancy rate of all dispatched tokens.}
    \label{fig:redundancy-rates}
    \Description{ }
\end{figure}

This problem is exacerbated in expert-specialized MoEs, which rely on large top-$k$ routing where each token is sent to a large number of experts.
If multiple selected experts reside on the same node, existing systems send multiple copies of the same token activation across inter-node links, one for each destination expert, even though only one copy is actually needed (\sref{subsec:rbd}).
To quantify this redundancy, we evaluate a DeepSeek-style configuration (256 experts, $k$=8 routing) using DeepSpeed-MoE. \fref{fig:redundancy-rates} shows that the redundancy rate ranges can be up to 75.1\%, depending on the EP size. This leads to redundant inter-node communication.

\smallskip
\noindent \fbox{\parbox{0.96\linewidth}{
\textbf{Takeaway-3:} Expert-specialized MoEs increase the number of routed expert per token, leading to significant duplication in communication. On HPC platforms with hierarchical and heterogeneous networks, this results in inefficient use of inter-node bandwidth and causes communication to become a major training efficiency bottleneck as the expert granularity increases.}}

\section{\name Design}
\label{sec:design}
\begin{figure*}[!ht]
    \centering
    \includegraphics[width=0.88\linewidth]{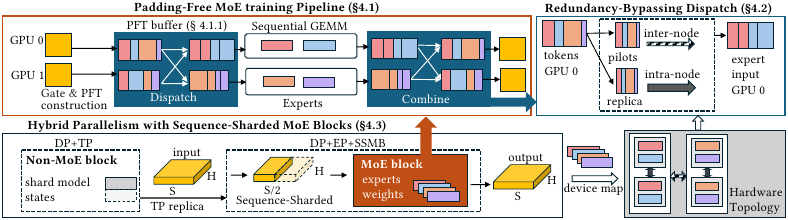}
    \caption{Overview of \name. At a high level, \name enables efficient and scalable training for expert-specialized MoEs through a set of targeted system-level optimizations, including a fully padding-free MoE training pipeline with cross-platform kernels (\sref{subsec:kernels}), redundancy-bypassing dispatch (\sref{subsec:rbd}), and hybrid parallelism with sequence-sharded MoE blocks (\sref{subsec:tsp}).}
    \label{fig:overview} 
    \Description{}
\end{figure*}

To address the training inefficiencies of emerging MoE architectures on non-NVIDIA platforms, \name introduces a set of system optimization techniques to address the unique challenges posed by fine-grained experts with large top-$k$ routing on hierarchical HPC platforms. \fref{fig:overview} provides an overview of \name. First, we propose a new sparse data layout PFT and rework the MoE pipeline to eliminate zero-padding across different MoE stages via padding-free sparse MoE training with cross-platform kernels (\sref{subsec:kernels}). Second, we reduce communication redundancy by leveraging topological awareness of HPC systems via a hierarchical two-stage redundancy-bypassing dispatch algorithm (\sref{subsec:rbd}). Third, \name incorporates a hybrid prallelism strategy that enables sequence-sharding in MoE blocks to address the activation memory bottleneck, with topology-aware planning and device-mapping. Together, these optimizations form an integrated and cross-platform training system that scales emerging MoEs on large scale HPC clusters. The following sections provide an in-depth description of each design. 

\subsection{Padding-Free Sparse MoE Training with Cross-Platform Kernels}
\label{subsec:kernels}

We introduce the truly padding-free MoE training pipeline in X-MoE. First, we design a novel sparse data-structure: PFT (\textbf{P}adding-\textbf{F}ree \textbf{T}oken buffers). The PFT is designed to eliminate zero padding through the MoE computation and communication stages, including the dispatch, MLP, and combine stages (\sref{subsubsec:padding-free-storage}). Instead of allocating fixed-capacity expert buffers padded with zeroed-vectors, PFT stores only valid routed tokens. 
However, in introducing the PFT to MoE training, each of the dispatch, MLP, and combine stages needs to be modified to operate on the PFT. We also detail our modifications to each stage (\sref{subsubsec:padding-free-storage}). 
Second, to efficiently implement the PFT-based pipeline, we design a suite of Triton-based kernels to handle the corresponding sparse and irregular workloads (\sref{subsec:pipeline-backend}). These kernels are designed to be hardware-agnostic, support coalesced memory accesses and avoid vendor-specific constraints like CUDA-only fused kernels. As a result, our padding-free MoE training pipeline improves memory efficiency and reduces communication volume, which serve as key enablers of scalable training of expert-specialized MoEs on diverse hardware. 

\subsubsection{Padding-Free Token Storage and Pipeline}
\label{subsubsec:padding-free-storage}
\begin{figure}[!ht]
  \centering
  \begin{minipage}[b]{0.45\textwidth}
    \centering
    \includegraphics[width=1\linewidth]{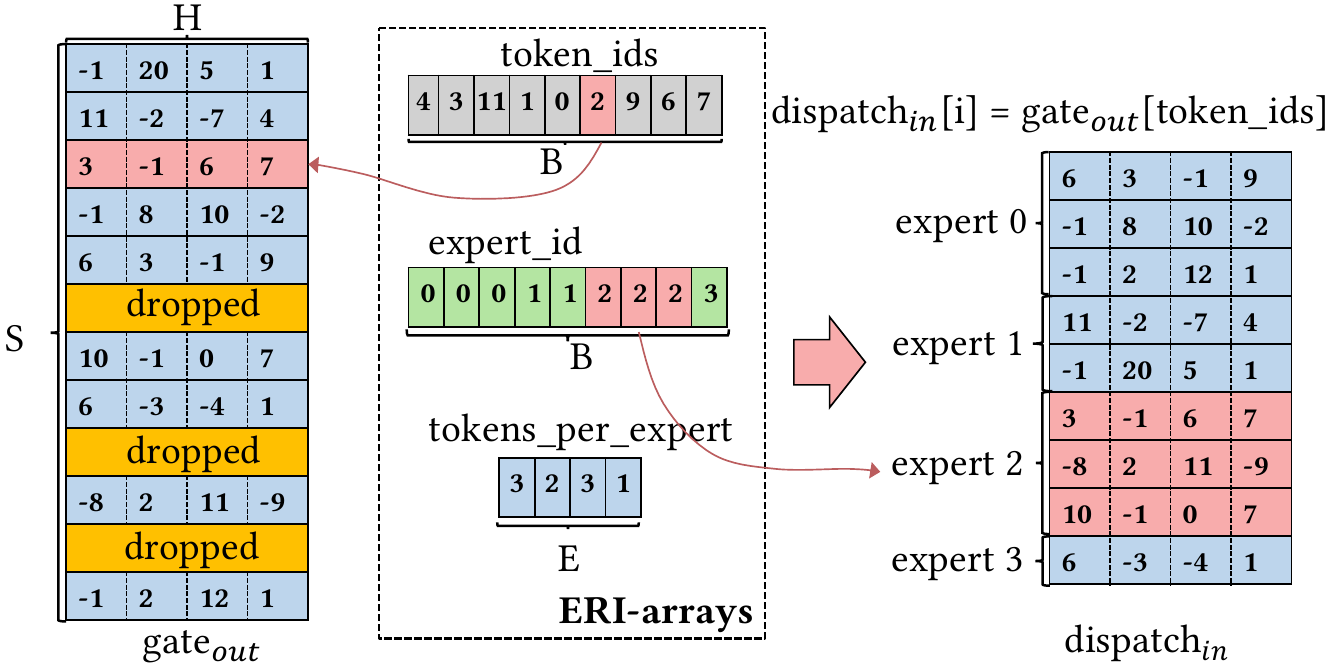}
    
    \caption{PFT with sparse structure and ERI-arrays.}
    \label{fig:pft}
    \Description{}
  \end{minipage}
  \hfill
\end{figure}

To eliminate the inefficiencies introduced by zero-padding in existing MoE pipelines, we introduce the PFT data-structure. Unlike standard expert input buffers that reserve fixed-capacity slots per expert, PFT consists of a token-buffer, \texttt{x}, which stores only the routed tokens, along with expert routing information arrays (\metadata) that track how each token should be processed. 
The \metadata consists of the following data: (1) array \texttt{token\_ids} (a \texttt{[B]}-sized array; \texttt{B} is the number of routed tokens in \texttt{x}) where \texttt{ti = token\_ids[i]} is an index that maps the \texttt{ti$^{th}$} input-token to the \texttt{i$^{\text{th}}$} position in the dispatch matrix (see figure \fref{fig:pft} for an example), 
(2) array \texttt{expert\_ids} (a \texttt{[B]}-sized array) where \texttt{expert\_ids[i]} represents the expert that \texttt{x[i]} is routed to, (3) array \texttt{tokens\_per\_expert} (a \texttt{[E]}-sized array; \texttt{E} is the expert count) where \texttt{tokens\_per\_expert[i]} represents the number of tokens in \texttt{x} routed to expert \texttt{i}, (4) array \texttt{combine\_weights} (a \texttt{[B]}-sized array) where \texttt{combine\_weights[i]} represents the value that \texttt{combine$_{\text{in}}$[i]} (an intermediate matrix assembled after the last \texttt{alltoall}, described later in~\sref{subsec:pft-construction}) is scaled by in the combine phase. We first show how PFT is constructed and then demonstrate how this representation allows each MoE stage to operate without any zero padding. \fref{fig:pft} depicts the PFT structure with \metadata and how \metadata drive token dispatching. 

\noindent
\textbf{\emph{PFT Construction.}}
\label{subsec:pft-construction}
PFT is constructed after the MoE gating function and before token dispatching (we describe how the gating, dispatch, MLP and combine stages are modified later). Listing~\ref{alg:pft-construction} illustrates the pseudo code for PFT construction. It takes as input the outputs of the gating function: (1) the \texttt{top\_experts} array, a \texttt{[S, E]}-sized array that contains the token to expert mapping and (2) corresponding \texttt{combine\_weights} array, a \texttt{[S, E]}-sized array used in the combine stage containing each token's probability score that reflects the confidence of the gating function. It also takes as input the \texttt{max\_token\_count} variable, indicating the expert-capacity. The PFT construction routine returns a PFT whose \metadata are correctly instantiated. 

\lstdefinestyle{pft-style}{
    language=Python,
    frame=single,
    framerule=0pt,
    backgroundcolor=\color{codebg},   
    commentstyle=\color{codepurple},
    keywordstyle=\color{codeblue},
    emphstyle=\ttfamily\color{codered},
    stringstyle=\color{codegreen},
    basicstyle=\ttfamily\linespread{0.85}\footnotesize,
    breakatwhitespace=false,         
    breaklines=true,                 
    captionpos=b,                    
    keepspaces=true,                 
    numbers=left,                    
    numbersep=-4pt,                  
    showspaces=false,                
    showstringspaces=false,
    showtabs=false,                  
    tabsize=2,
    morekeywords={}
}

\lstset{
  mathescape,         
  literate={->}{$\rightarrow$}{2}
           {ε}{$\varepsilon$}{1}
}

\begin{lstlisting}[style={pft-style}, 
    label={alg:pft-construction},
    caption={Padding-free MoE-layer},
    float=!ht]
  def gating(k,tokens):
    """
    k: int.
    tokens: input tokens to MoE-layer [S, H]-sized.
    """
    logits = softmax(FFN(tokens), axis=-1)
    combine_weights, top_experts = topk(logits)
    return top_expert, combine_weights, tokens
    
  def PFT_construction(max_token_count,top_experts,combine_weights):
    """
    top_experts: token to expert assignments [S, E]-sized
    combine_weights: gating probabilities [S, E]-sized 
    max_token_count: maximum tokens for a single expert
    Returns generated PFT
    """
    E = get_expert_count()
    ## Step 1: Generate the expert & token ids ##
    # shape (lines 20-21): [S*K]
    flat_top_experts = flatten(top_experts)
    expert_ids, token_ids = sort(flat_top_experts)
    ## Step 2: Identify filter dropped-tokens ##
    # shape (lines 24-26): [S*K]
    flat_combine_weights = flatten(combine_weights)
    sorted_indices = argsort(flat_combine_weights)
    sorted_top_experts = flat_top_experts[sorted_indices]
    # shape (lines 28-30): [S*K, E] 
    one_hot_enc=one_hot(sorted_top_experts,num_classes=E) 
    rank_in_expert = cumsum(one_hot_enc, axis=0)  
    weight_mask = rank_in_expert <= max_token_count
    # shape (lines 32-36): [B], token-count post dropping
    filtered_indices = sorted_indices[weight_mask]
    retained_token_ids = isin(flat_top_experts, filtered_indices)
    token_ids = token_ids[retained_token_ids]
    expert_ids = expert_ids[retained_token_ids]
    combine_weights = combine_weights[retained_token_ids]
    # shape: [E]
    tokens_per_expert = histogram(expert_ids, bins=E)  
    # Returns a PFT with generated ERI-arrays
    return PFT(token_ids,expert_ids,tokens_per_expert,combine_weights)
 
  def dispatch(pft, gate$_{out}$):
    dispatch$_{in}$ = gather_kernel(gate$_{out}$,pft.token_ids,pft.expert_ids)
    pft.tokens_per_expert=alltoall(pft.tokens_per_expert)
    dispatch$_{out}$ = alltoallv(dispatch$_{in}$, pft.tokens_per_expert)
    pft.x = dispatch$_{out}$
    return pft

  def mlp(pft, w1, w2):
    inter_activ = sequential_gemm(pft.x,w1)
    mlp$_{out}$ = sequential_gemm(inter_activ,w2)
    pft.x = mlp$_{out}$
    return pft

  def combine(pft):
    combine$_{in}$ = alltoallv(pft.x,pft.tokens_per_expert)
    combine$_{out}$ = scatter_kernel(combine$_{in}$,pft.token_ids,pft.expert_ids,pft.combine_weights)
    return combine$_{out}$

  def call(tokens, k, max_token_count, w1, w2):
    """
    tokens: tokens input to the MoE layer, [S, H]-sized
    k: topk value, int.
    max_token_count: expert capacity, int.
    w1 & w2: weights of first and second layer of  MLP.
    """
    top_expert,combine_weights,gate$_{out}$ = gating(k,tokens)
    pft = PFT_construction(max_token_count,top_expert,combine_weights) 
    pft = dispatch(pft,gate$_{out}$)
    pft = mlp(pft,w1,w2)
    pft = combine(pft)
    return pft.x\end{lstlisting}

The PFT construction routine proceeds in two stages. In the first stage, we flatten and sort the incoming \texttt{top\_experts} array (lines 20-21), which contains the token to expert assignments generated by the gating function. In the second stage, we determine which tokens are dropped (lines 24-33); using this information, we construct the \metadata by pruning out the dropped tokens from the unfiltered \texttt{token\_ids}, \texttt{expert\_ids}, and \texttt{combine\_weights} (lines 34-36). 

\noindent
\textbf{\emph{Padding-free Gating, Dispatch, MLP and Combine.}} Listing ~\ref{alg:pft-construction} (lines 67-72) illustrate our padding-free pipeline where the modified dispatch, MLP and combine stages operate on the PFT. First, during gating, we transform the input tokens to logits and select the top-k experts per token, returning their respective expert indices (\texttt{top\_expert}) and probability confidence scores of the assignment (\texttt{combine\_weights}) (lines 6-8). Second, we construct the PFT structure using these data. Third, during dispatch, we consume the pft and \texttt{gate$_{\text{out}}$} tokens and route tokens to the correct worker. This occurs by: (1) reordering the tokens locally using a custom gather-kernel (described in \sref{subsec:pipeline-backend}) producing the \texttt{dispatch$_{\text{in}}$} buffer, (2) exchanging the tokens between devices via an uneven \texttt{alltoall}, routing each token to the correct device its expert resides on (lines 43-47) producing the \texttt{dispatch$_{\text{out}}$} buffer. No zero-padding is communicated in this stage. Fourth, the MLP layer processes tokens; we launch a custom sequential-GeMM (described in \sref{subsec:pipeline-backend}) to implement each MLP layer enabling different token-counts to be multiplied by the MLP weights of different experts without the need for zero-padding (lines 50-53). Fifth, during combine, tokens are re-routed back to their original device and a custom scatter kernel (described in \sref{subsec:pipeline-backend}) locally reorders the inbound tokens to their original position in the sequence, multiplying each token with its respective value in \texttt{combine\_weights} (lines 56-58). 

\subsubsection{Highly-Optimized Cross-Platform Sparse and Irregular Kernels}
\label{subsec:pipeline-backend}

The PFT format helps improve the memory-efficiency of emerging MoE training by eliminating the need for any zero-padding in the dispatch, MLP and combine stages. However, certain operators in this modified pipeline, such as the gather, scatter and sequential GeMM, introduce sparse and irregular access patterns to the PFT \metadata, which can be expensive to implement in Pytorch and requires specialized kernels for efficiency.
To address these issues, we introduce Triton-based gather, scatter as well as (non Triton-based) sequential GeMM implementations that are high-performance and platform agnostic. The gather and scatter kernels are responsible for computing: \texttt{dispatch$_{\text{in}}[i, :]$ = gate$_{\text{out}}[token\_ids[i], :]$} and \newline \texttt{combine$_{\text{in}}$[token\_ids[i], :] = mlp$_{\text{out}}$[i, :] \newline $\times$ \texttt{combine$\_{\text{weights}}$}[token\_ids[i], :]}, respectively. However, the irregular memory access patterns in reading and writing to tensors results in uncoalesced memory requests and poses a unique performance challenge. We circumvent this by scheduling a single thread-block to operate (read and write) on one vector, assigning contiguous threads to operate on the model-hidden dimension (outer-dimensions of the \texttt{gate$_{\text{out}}$ and \texttt{combine$_{\text{in}}$} tensors}). On the other hand, our sequential GeMM operates on the \texttt{dispatch$_{\text{out}}$} matrix. It extracts the correct tokens each expert is assigned to in \texttt{dispatch$_{\text{out}}$} with a python for-loop launching \texttt{E$_{\text{local}}$} (number of experts assigned to the device) GeMMs. 

\subsection{Redundancy-Bypassing Dispatch}
\label{subsec:rbd}

\begin{figure}[t]
    \centering
    \includegraphics[width=1\linewidth]{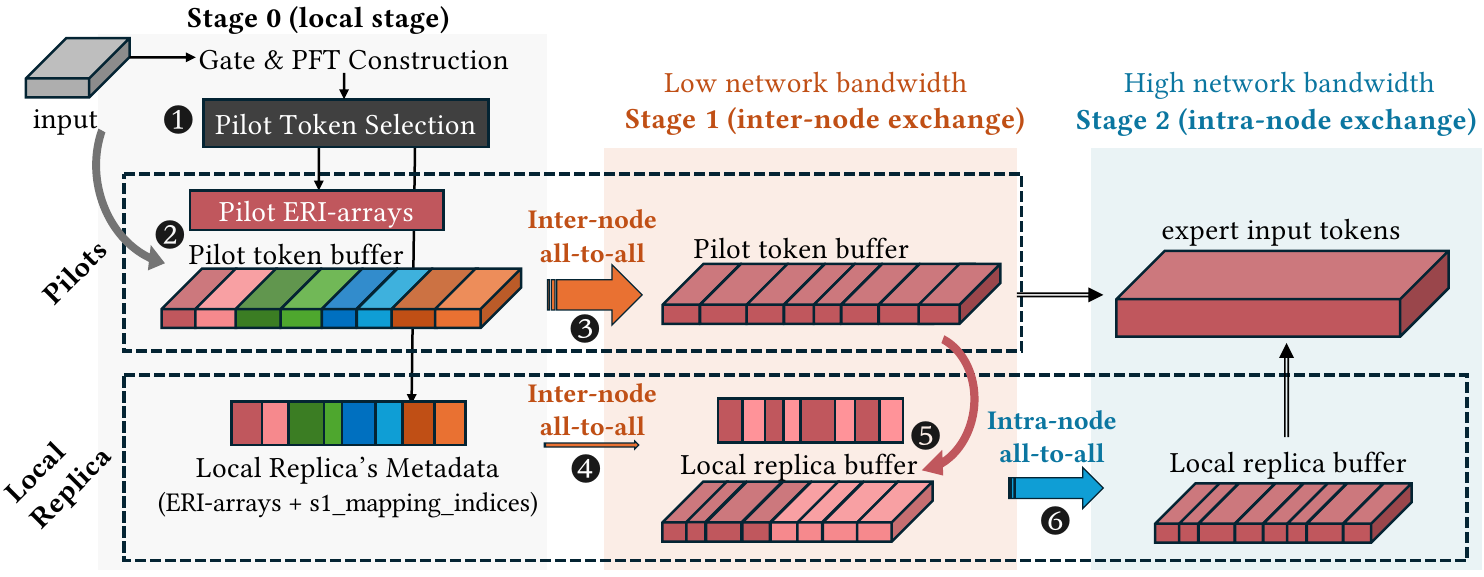}
    \caption{Hierarchical Redundancy-Bypassing Dispatch: Multi-stage token routing across inter- and intra-node networks to reduce communication duplication.}
    \label{fig:redundancy-bypassing-impl}
    \Description{}
\end{figure}

We propose Hierarchical Redundancy-Bypassing Dispatch (\rbd) to eliminate redundant inter-node communication by introducing a multi-stage dispatching process with two groups of tokens: \emph{Pilot tokens}, which are the minimal set of distinct tokens that must be communicated across nodes; and \emph{local replica}, which are local duplicates of pilot tokens routed to additional experts on the same destination node. 
Instead of sending all token data through one \texttt{alltoall}, \rbd only sends pilot tokens through inter-node communication and propagates local replica using fast intra-node connects.  
We now illustrate \rbd's multi-stage process using \fref{fig:redundancy-bypassing-impl}. 

\noindent
\textbf{\emph{Stage 0 (S0)}: Pilot Selection and Instantiation.} The first step of \rbd is \emph{pilot tokens selection} within $x$, which is generated through PFT in \sref{subsec:kernels}.
Based on \texttt{token\_ids} and \texttt{expert\_ids}, \rbd extracts the node destination information for each token. Then for each token's $k$ destinations, \rbd identifies the group of experts that share the same destination node. Among tokens with the same source and destination node, \rbd randomly selects one as the pilot token and marks the rest as local replica. 
This randomized strategy helps avoid a biased distribution and creates a balanced workload for \texttt{alltoall} communication. For example, always routing tokens to the smallest expert ID within a node will significantly increase the \texttt{alltoall} latency.

Meanwhile, we create separate \metadata for pilot tokens and local replicas, respectively. Each of them contains the routing information for those tokens. This process is represented by \ding{182} in \fref{fig:redundancy-bypassing-impl}. In addition, we construct a mapping array \texttt{s1\_mapping\_indices} (used in Stage 1 for local replica reconstruction), where each local replica token records the index of its corresponding pilot token. 
To ensure the correctness of this mapping index before and after the uneven \texttt{alltoall} exchange (\ding{185}), we use the relative index starting from 0 for each target expert. This is allowed because the pilot \metadata is sorted by expert IDs. We re-encode it to the absolute index after the \texttt{alltoall} exchange.
Finally, we instantiate the pilot token buffer (\ding{183}) using a Triton gather kernel. The local replica tokens are not instantiated yet. They are reconstructed from their associated pilot tokens after the pilot tokens arrive their destination.

\noindent
\textbf{\emph{Stage 1 (S1)}: Inter-Node Exchange (Pilot Only) and Local Replica Reconstruction.} 
In S1, \rbd sends only pilot tokens across nodes using an uneven \texttt{alltoall} (\ding{184}). This is the only stage that uses inter-node bandwidth. Additionally, \rbd also sends local replica tokens' metadata (\metadata and \texttt{s1\_mapping\_indices}) (\ding{185}), alongside their corresponding pilot tokens. This is lightweight given that metadata has small message size. 
Once the pilot tokens arrive at their destination node, local replica tokens are reconstructed by copying data from pilot tokens to a local exchange buffer based on \texttt{s1\_mapping\_indices} (\ding{186}). Note that the local exchange buffer serves as the input of the intra-node \texttt{alltoall}, \rbd ensures token data is contiguous and ordered by destination (e.g., the ascending order of expert IDs).

\noindent
\textbf{\emph{Stage 2 (S2)}: Intra-Node Exchange (Local Replica Only) and Expert Input Reconstruction.} 
The newly reconstructed local replica tokens are exchanged among GPUs within the same node using a fast intra-node uneven \texttt{alltoall} \ding{187}, which helps to save expensive inter-node traffic. After pilot tokens and local replica tokens all arrive at their target GPUs, \rbd reconstructs the each expert's local input by merging the two groups and correctly orders them based on their expert indices.

The combine stage reverses the above described \rbd process. 
Specifically, local replica tokens are first gathered via intra-node communication, followed by pilot tokens through inter-node transfer. 
To ensure the correctness of combining weight scaling on expert outputs, we exchange the original \texttt{combine\_weights} \metadatum for all tokens in advance through a small inter-node \texttt{alltoall}, along with \ding{185}. During combine, the weight scaling is performed in \emph{stage 1}, before merging local replica tokens into pilot tokens. Finally, the full results are reconstructed from the pilot tokens on the original device using the \metadata preserved during dispatch.

\subsection{Hybrid Parallelism with Sequence-Sharded MoE Blocks}
\label{subsec:tsp}

Training MoEs at scale requires hybrid parallelism that carefully balances memory, compute, and communication.
A common approach is to apply tensor parallelism (TP) to dense blocks (e.g., attention, non-MoE MLPs) as in Megatron-LM~\cite{megatron-lm} and switch to expert-parallelism (EP) for MoE blocks, sharding expert weights across devices. This TP + EP combination, used in systems such as DeepSpeed-TED~\cite{deepspeed-ted}, allows scaling conventional MoE parameters across large clusters. However, the naive transition from TP to EP fails to address the key bottleneck in expert-specialized MoEs: the activation memory, especially for \adispatch and \acombine. As described in \sref{subsec:bottleneck-shift}, these tensors scale linearly with sequence length $s$, routing factor $k$, hidden dimension $h$, and fine-grained factor $m$.

Tensor parallelism works by duplicating input tokens across all TP ranks and computing partial results, which are later reduced via all-reduce. In the context of MoE training, this means that each TP worker holds a full copy of the input sequence, and even if we switch to EP within the MoE block, each EP worker begins the MoE computation with the same \textit{duplicated activations}. Consequently, the heavy activations (\adispatch and \acombine as described in \sref{subsec:bottleneck-shift}) are not reduced at all, as they are still duplicated across all EP workers that originate from TP ranks, which leads to poor scaling.

To address this challenge, we propose a new hybrid parallelism strategy that combines tensor-slicing parallelism with sequence-sharded execution for MoE Block (\ssmb).
This strategy is motivated by a key insight: 
all operations in an MoE block (gating, dispatch, expert FNNs, and combine) are applied token-wise and do not require inter-token dependencies. This allows us to shard the input sequence of the MoE block across EP ranks, so each rank only retains and processes a segment of the sequence and later recover the full sequence using all-gather, reintroducing the duplicated inputs expected by the next TP block. This strategy reduces the activation footprint of \adispatch and \acombine by a factor of the TP group size, while preserving compatibility with standard MoE routing and communication. 

\begin{figure}[t]
    \centering
    \includegraphics[width=1\linewidth]{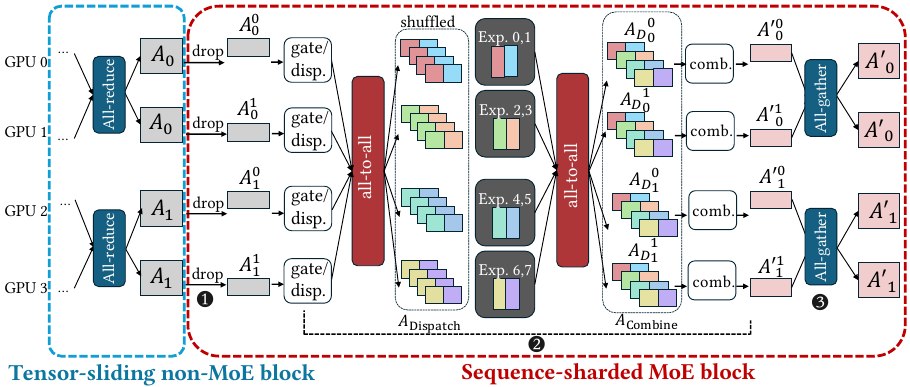}
    \caption{Illustration of \name's hybrid parallelism with sequence-sharded MoE blocks.}
    \label{fig:ssp}
    \Description{}
\end{figure}

\fref{fig:ssp} illustrates how \ssmb works in practice. In this setup, we use TP=2 and DP=2 (TP and DP parallel-group size) for the dense (non-MoE) block, and EP=4 for the MoE blocks. In the TP + DP phase, each TP worker holds a full copy of the input sequence: device 0 and 1 each have a copy of sequence $A_0$, while device 2 and 3 have $A_1$. In existing MoE training, duplicated activations like \adispatch and \acombine would be store on both devices, increasing memory cost. Instead, \ssmb drops a fraction of the tokens on each device (\ding{182}), partitioning the sequence across TP ranks (e.g., $A_0^0$ and $A_0^1$). After entering the MoE block, \ssmb reassigns each TP+DP worker to act as an EP rank and performs MoE-Gating, dispatch, expert FNNs, and combine on the partitioned tokens (\ding{183}), using the padding-free pipeline introduced in \sref{subsec:kernels}. After the combine op, \ssmb issues an all-gather (\ding{184}) to reconstruct the full output sequence (e.g., $A_0^\prime$) across all EP ranks, effectively restoring the replicated data layout for the next TP-based non-MoE block. 

In the backward pass, \ssmb follows a reversed sequence of operations. Upon entering the MoE block, it first drops the gradients corresponding to the partial sequences retained during forward. It then performs expert-specific gradient computation and \texttt{alltoall} communications, mirroring the forward process. Finally, \ssmb uses an all-gather operation to reconstruct the full input gradient across TP ranks, allowing the propagation to continue in the TP phase. 

\noindent
\textbf{\emph{Can existing parallel strategies handle the shifted memory bottleneck?}} The careful reader may think of an alternative approach that uses TP + EP for MoE blocks, as opposed to EP with sequence sharding. After all, these schemes also shard model states across devices and reduce memory. However, in expert-specialized MoEs, experts already have small intermediate dimensions, making TP's benefits marginal. Moreover, neither TP nor ZeRO-style DP reduce the expensive activations like \adispatch and \acombine. 
We compare the overhead and gain of using \ssmb with tensor-expert-data (TED) parallelism by calculating both model states and activation memory saving of each approach. In short, the benefit of \ssmb over TED depends on the ratio: $r=\frac{k}{H_{FFN}}$. Under the usage of ZeRO-1 DP, when this ratio satisfies: $r > \frac{2}{c\cdot S}$, \ssmb offers more memory savings than TED. For expert-specialized MoEs with fine-grained factor $m$, we have $H_{FFN}\propto \frac{1}{m}$ and $k\propto m$. Thus, generally speaking, under identical sequence length choice, the more fine-grained the MoE model is, the more benefits \ssmb provides over TED. 

\noindent
\textbf{\emph{Why not activation checkpointing?}} Another approach to reducing activation memory is activation checkpointing~\cite{activation-checkpoint}, which trades memory for recomputation. However, in MoE training with expert parallelism, \adispatch and \acombine are outputs of \texttt{alltoall} communication during token routing. 
In regular MoE training, 4 \texttt{alltoall}s are needed per layer in each step. SSMB follows this, requiring 4 \texttt{alltoall}s per step. However, applying checkpointing to these tensors would require two extra \texttt{alltoall} communications during the backward pass, resulting in a total of 6 \texttt{alltoall}s per layer, which incurs expensive communication overhead in addition to the recomputation overhead.

\noindent
\textbf{\emph{Why not use pipeline parallelism (PP)?}}  While PP is effective for reducing memory by splitting the model across devices, it requires significant code refactoring and careful scheduling to balance pipeline stages, especially with sparse MoE layers. In contrast, our solution requires minimal code changes. We leave the integration with PP as future work.

\section{Evaluation}
\label{sec:eval}

In this section, we evaluate \name in comparison with state-of-the-art large-scale MoE training approaches, demonstrating that it achieves significantly improved training efficiency and scalability for emerging MoEs. We also show the impact of different technologies within \name on performance.

\subsection{Evaluation Methodology}
\label{subsec:eval-method}

\noindent
\textbf{Hardware.} We conduct evaluation on the \emph{Frontier} supercomputer \cite{frontier}. Each cluster node is equipped with 4$\times$AMD MI250X GPUs with dual Graphics Compute Dies (GCDs) and one EPYC CPU. A GCD is viewed as an effective GPU. The 2 GCDs on the same MI250X are connected with Infinity Fabric with a peak bandwidth of 200 GB/s. The GCDs on different MI250X are connected with Infinity Fabric where the peak bandwidth ranges from 50-100 GB/s. The Frontier nodes are connected with four Slingshot 25 GB/s NICs. We use up to 128 nodes (1024 MI250X GCDs) for experiments. 

\noindent
\textbf{Evaluation setup.} We implement \name in DeepSpeed~\cite{deepspeed}, a widely used open-source DL training library. We include the implementation and environment details in Appendix. If not specified, we choose the maximum micro-batch size of power of 2 under the memory limitation and a global batch size of 1024. We choose the capacity factor $c=1.25$ for all experiments, as suggested by \cite{g-shard}.

\noindent
\subsection{Main Results}
\label{subsec:main-results}

We first demonstrate that \name scales effectively across a wide range of expert-specialized MoE models.
We use model configurations from DeepSeek-MoE~\cite{deepseek-moe}, DeepSeek-v2~\cite{deepseek-v2}, and DeepSeek-v3~\cite{deepseek-v3}, as shown in Table~\ref{tbl:model-config}. We compare \name against three large-scale MoE training frameworks: DeepSpeed-MoE~\cite{deepspeed-moe}, DeepSpeed-TED~\cite{deepspeed-ted}, and Tutel~\cite{hwang2023tutel} as baselines. For DeepSpeed-MoE and Tutel, we sweep EP size in \{32/64/128/256\} and ZeRO stages 1/2. For DeepSpeed-TED, we additionally sweep TP in \{1, 2, 4, 8\} and choose the best performing configuration. 

\begin{table}[!ht]
\centering
\small
\begin{tabular}{l|ccc|c}
\toprule
\textbf{Models} & \textbf{Small} & \textbf{Medium} & \textbf{Large} & \textbf{Super}\\
\midrule
seq. length         & 2048 & 4096 & 4096 & 4096                   \\
$H_{model}$         & 2048 & 5120 & 7168 & 7168                   \\
$H_{FFN}$           & 1408 & 1536 & 2048 & 2560                   \\
num. experts          & 64   & 128  & 256 & 256                       \\
top-$k$                 & 6    & 6    & 8 & 8                   \\
num. layers           & 28   & 28   & 28  & 61     \\
\midrule
Param.                 & 10.1 B & 55.2 B & 201.4 B & 545.4 B                   \\
Activated Param.                 & 1.3 B & 5.2 B & 11.5 B & 28.7 B                     \\
\bottomrule
\end{tabular}
\caption{The model configs used for evaluation.}
\label{tbl:model-config}
\end{table}

\begin{figure}[!ht]
    \centering
    \includegraphics[width=1\linewidth]{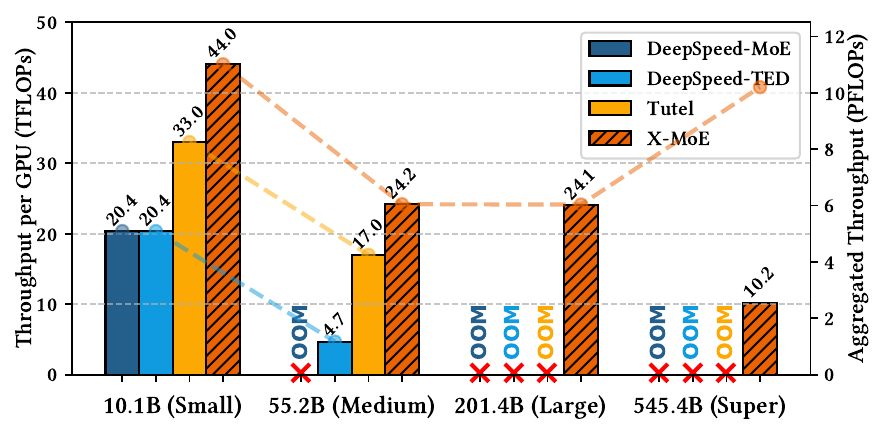}
    \caption{Results on training Small, Medium, and Large models on 256 GPUs; training Super model on 1024 GPUs. The dashed lines show the aggregated throughput.}
    \label{fig:scale-model-sizes} 
    \Description{}
\end{figure}

\noindent
\textbf{\emph{Trainability and throughput.}}
We evaluate \name on Small (10.1B), Medium (55.2B), and Large (201B) model configurations using 256 GPUs. 
As shown in \fref{fig:scale-model-sizes}, while existing systems such as DeepSpeed-MoE, DeepSpeed-TED, and Tutel run out of memory on medium and large models, \name successfully enables their training, effectively changing the status from non-trainable to trainable under the same hardware budget. When multiple systems can train a model, e.g., on the \emph{Medium} model, \name achieves higher throughput, with 5.15x and 1.42x speedup over DeepSpeed-TED and Tutel respectively. 
\name achieves these results through a set of targeted system-level optimizations. Its padding-free training pipeline eliminates zero-padding overhead in both memory and communication. \rbd reduces communication redundancy in high top-$k$ routing scenarios by minimizing cross-node token duplication. 
\ssmb effectively mitigates the shifted memory bottleneck. Together, these innovations enable efficient and scalable training of emerging expert-specialized MoEs. 

\noindent
\textbf{\emph{Pushing the model scale limit.}} \name further enables the \emph{Super} 545B model on 1024 GPUs, achieving an aggregated throughput of 10.44 PetaFLOPs while all prior systems fail due to OOM errors. At this scale, training becomes sensitive to system dynamics beyond memory and communication volume optimizations. 
On Frontier, we observe that scaling beyond 256 GPUs results in significantly higher \texttt{alltoall} latencies ($>10\times$ higher than average), likely due to increased cross-rack communication and network congestion from concurrently running jobs on the shared cluster.
Despite this, \name successfully sustains large-scale MoE training across racks, demonstrating its robustness and extending the boundary of what is trainable on today's HPC clusters.

\subsection{Scalability Evaluation}
\label{subsec:eval-scalability}

We evaluate both weak and strong scaling to demonstrate that \name not only enables training of expert-specialized MoEs but also scales efficiently with increasing compute resources. We compare against only Tutel, as it is the best performing baseline as shown in \fref{fig:scale-model-sizes}.

\begin{figure}[!ht]
    \centering
    \begin{minipage}{0.48\linewidth}
        \centering
        \subfloat[\label{fig:weak-scaling}]{
            \includegraphics[width=1.03\linewidth]{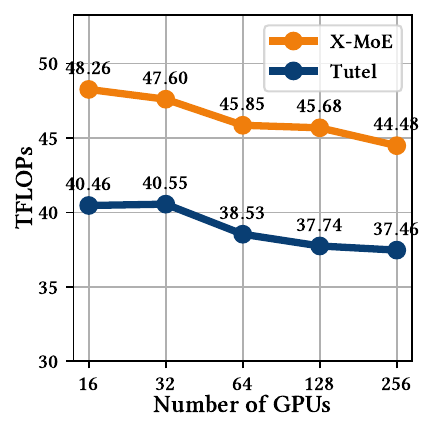}
        }
    \end{minipage}
    \hfill
    \begin{minipage}{0.48\linewidth}
        \centering
        \subfloat[\label{fig:strong-scaling}]{
            \includegraphics[width=1.03\linewidth]{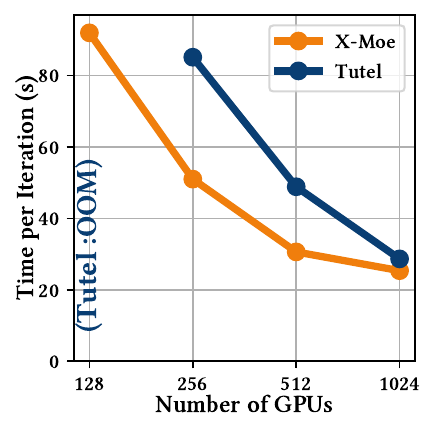}
        }
    \end{minipage}
    \caption{Scalability results. (a) Weak scaling results: Training the 10.1B MoE on 16--256 GPUs with increasing batch size. (b) Strong scaling result: Training the 55.2B MoE across 128, 256, 512, 1024 GPUs while fixing the batch size.}
    \label{fig:scalibility}
    \Description{Weak and strong scaling results across different GPU counts.}
\end{figure}

\noindent
\textbf{\emph{Weak scaling.}} To evaluate weak scaling behavior, we train the 10.1B \emph{Small} model from 16 to 256 GPUs, proportionally increasing the global batch size from 256 to 4096. We use EP=8 and scale out via ZeRO-DP. Our results are in figure \fref{fig:scalibility}(a).
The results show that \name consistently maintains higher TFLOPs compared to Tutel with a comparatively smaller drop in throughput as the number of GPUs increases. 


\noindent
\textbf{\emph{Strong scaling.}} We evaluate strong scaling using the 55.2B \emph{Medium} model on 128, 256, 512, and 1024 GPUs, keeping the global batch size fixed at 2048. This setup tests how well the system reduces iteration time as more GPUs are used. Since DeepSpeed-MoE fails to run due to OOM errors, we compare \name (EP=64) with Tutel (EP=128). \fref{fig:strong-scaling} shows that Tutel cannot run on 128 GPUs even with EP=128, while \name scales effectively and achieves lower iteration time as GPU count grows. At 1024 GPUs, both systems converge to similar performance, as the increasing \texttt{alltoall} latency (as in \sref{subsec:main-results}) becomes the dominant bottleneck at this scale. 


\noindent
\subsection{Analysis Results}
\label{subsec:eval-breakdown}

\subsubsection{How does PFT and the padding-free pipeline bring benefits?} We evaluate the benefits of PFT format and the associated padding-free pipeline in two dimensions: (1) reduced layer-wise execution time, and (2) improved memory efficiency. 

\begin{figure}[t]
    \centering
    \includegraphics[width=1\linewidth]{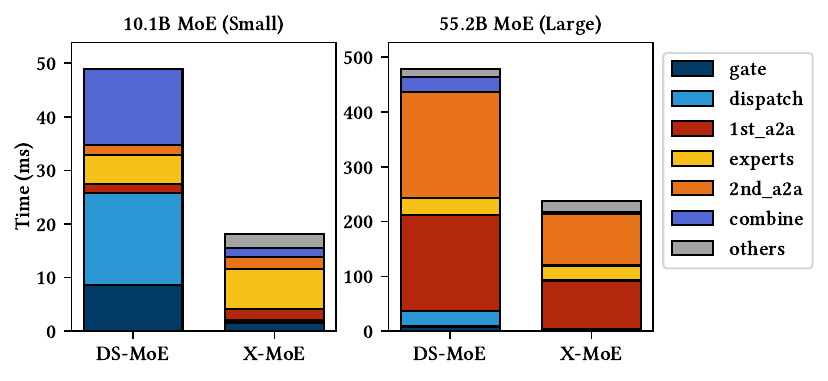}
    \caption{Forward MoE layer time breakdown comparison between DeepSpeed-MoE and \name of training \emph{Small} model and \emph{Large} Model.}
    \label{fig:ptf-breakdown}
    \Description{}
\end{figure}

\noindent
\textbf{\emph{MoE layer time breakdown.}} To evaluate the impact of PFT, we compare the MoE layer time breakdown of \name and DeepSpeed-MoE when training the \emph{Small} model (EP=8) and \emph{Large} model (EP=64) on 256 GPUs. We disable other optimizations such as \rbd to isolate the contribution of PFT. \fref{fig:ptf-breakdown} shows the time comparison in an MoE layer: gating, buffer dispatching, dispatch \texttt{alltoall}, expert computation, combine \texttt{alltoall}, and buffer combining. 

The latency reduction arises for different reasons in the \emph{Small} and \emph{Large} models. 
For the \emph{Small} model, a major inefficiency of the baseline comes from the inefficient gating and buffer dispatch/buffer combine. \name improves on these stages due to the PFT sparse structure and efficient Triton kernels. Specifically, the gating, buffer dispatch and buffer combine stages are accelerated by 5.7$\times$, 35.7$\times$ and 8.1$\times$ respectively. Note that the expert computation time is slightly increased in \name at this scale. The reason is that \name applies a sequential GEMM on the uneven token buffer, which requires extra data transformations to get the expert input. Despite this overhead, the overall layer time is reduced by 62.3\%.
For the \emph{Large} model, the largest latency reduction comes from the \texttt{alltoall}. \name significantly reduces this time by 50.7\% by eliminating zero-padding. The gating, buffer dispatch and combine time are also negligible after \name's optimizations.

\noindent
\textbf{\emph{Activation memory savings.}}
We compare the per-layer activation memory usage of DeepSpeed-MoE, Tutel, and \name when training the \emph{Large} model on 256 GPUs, using EP=64 and ZeRO-style data parallelism. We report the maximum memory usage across all ranks. As shown in Table~\ref{tbl:pft-activation-memory}, \name achieves significantly lower memory consumption than both DeepSpeed-MoE and Tutel, because it reduces memory wastage on dispatching metadata as well as the unused tokens through its padding-free pipeline. Besides, another reason for Tutel's high memory usage is that the Tutel kernel forces the use of float32 on \acombine on AMD GPUs.

\begin{table}[!ht]
    \centering
    \begin{tabular}{c|ccc|c}
        \toprule
        & DS-MoE & Tutel & \name & Theoretical \\
        \midrule
        Memory (GB) & 2.81 & 1.95 & 1.21 & 1.125  \\
        \bottomrule
    \end{tabular}
    \caption{The activation memory consumption per-MoE-layer.}
    \label{tbl:pft-activation-memory}
\end{table}

\subsubsection{How does \rbd reduce MoE dispatching latency?}
\begin{figure}[t]
    \centering
    \includegraphics[width=1\linewidth]{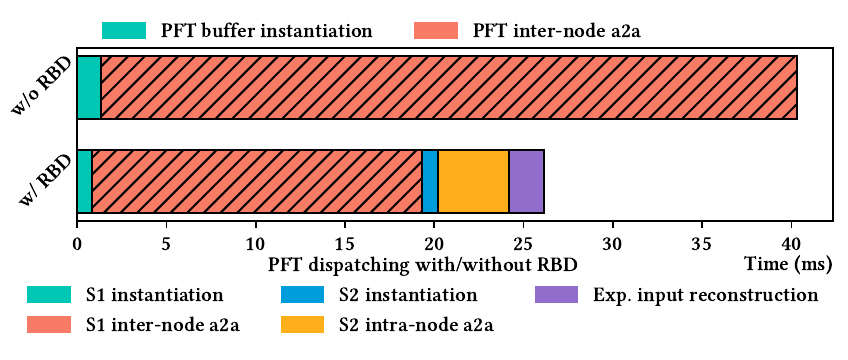}
    \caption{Dispatching time breakdown: With and without \rbd under PFT-based dispatching.}
    \label{fig:rbd-breakdown}
    \Description{}
\end{figure}

We evaluate the performance impact of dispatching with and without \rbd, with PFT format and padding-free pipeline enabled. We conduct the experiment using a single MoE layer from the \emph{Large} model on 32 GPUs with EP=32. In this setting, the measured redundancy rate is 54.8\%. \fref{fig:rbd-breakdown} shows that inter-node \texttt{alltoall} communication (shadowed area) dominates the total dispatch time in the padding-free training pipeline. \rbd reduces the inter-node communication time by 52.5\% by bypassing redundant tokens transferred through the low-bandwidth inter-node links. Although \rbd introduces extra overhead, such as intra-node \texttt{alltoall} (yellow) and data transformation costs, they are relatively minor compared to the savings from reduced inter-node data transfer volume, resulting in an overall performance speedup of 1.55x.

\noindent
\subsubsection{How does \ssmb save memory?} 
We evaluate the memory saving benefits of \name by comparing \name with \ssmb enabled and \name that uses the conventional tensor-expert-data parallelism (TP+EP+DP) without sequence sharding in MoE blocks, on the \emph{Large} model across 256 GPUs.
We enable ZeRO-1 DP and set EP=64 while varying the TP degree from 1 to 4. \fref{fig:ssmb} shows that enabling \ssmb leads to significantly lower memory usage, and the benefit grows as the TP degree increases. This is because \ssmb shards sequences within MoE blocks, effectively addressing the shifted memory bottleneck in expert-specialized MoEs. As model size increases, the TP degree naturally grows for non-MoE blocks, making \ssmb increasingly important for high memory efficiency for MoE training at scale. 

\noindent
\subsubsection{How does \ssmb compare to activation checkpointing?} 
\begin{figure}[!ht]
 \centering
 \begin{minipage}[c]{0.25\textwidth}
  {\includegraphics[scale=0.68]{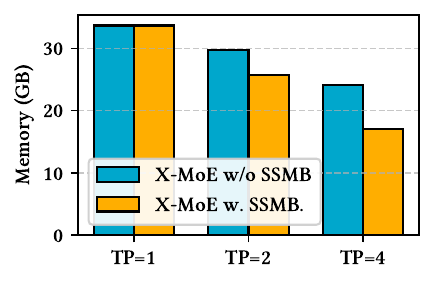}}
  \caption{\name's maximum allocated memory across GPUs
w/ and w/o SSMB.}\label{fig:ssmb}
 \end{minipage}%
 \hfill
 \begin{minipage}[c]{0.2\textwidth}
  {\includegraphics[scale=0.68]{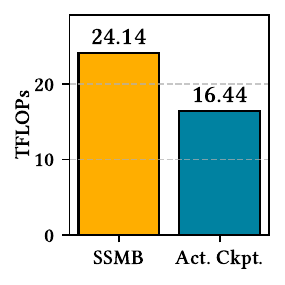}}
  \caption{TFLOPs of enabling SSMB vs. activation checkpointing.}\label{fig:ssmb-vs-activation}
  \Description{}
 \end{minipage}%
\end{figure}

One may ask how \ssmb compares to activation checkpointing, a technique that reduces memory usage. As shown in \fref{fig:ssmb-vs-activation}, under similar memory savings, \name with \ssmb achieves higher throughput. This is because \ssmb reduces activation memory without the cost of recomputation and extra \texttt{alltoall} during backward pass.

\subsection{Cross-platform Performance}

\begin{table}[!ht]
\centering
\begin{tabular}{lccc}
        \toprule
\textbf{TFLOPs} & \textbf{DeepSpeed-MoE} & \textbf{Tutel} & \textbf{X-MoE} \\
\midrule
Small (s=2048, l=28) & OOM   & OOM   & 46.87 \\
\midrule
Small-SR (s=1024, l=28) & 27.08 & 28.26 & 27.33 \\
Small-LR (s=2048, l=14) & 52.15 & 64.00 & 62.51 \\
\bottomrule
\end{tabular}
\caption{TFLOPs comparison of DeepSpeed-MoE, Tutel and X-MoE on 8$\times$NVIDIA A100 40GB GPUs. The Small model is the 10.1 B model listed in Table~\ref{tbl:model-config}. 
"Small-SR" and "Small-LR" models refer to sequence length reduced (SR) or number of layers reduced (LR) while maintaining the other configurations unchanged.}
\label{tbl:nvidia-eval}
\end{table}
To show X-MoE’s portability and performance beyond the AMD GPU platform, we evaluate X-MoE on eight NVIDIA A100 40 GB GPUs and compare against DeepSpeed-MoE and Tutel (Table \ref{tbl:nvidia-eval}). We train the 10.1 B (Small in Table \ref{tbl:model-config}) model in this experiment. Under the full 2k sequence length and 28 layers, both competing frameworks encounter out-of-memory (OOM) failures, whereas X-MoE sustains training at 46.87 TFLOPS. To further show the throughput comparison, we reduce either the sequence length to 1k (Small-SR) or the depth to 14 layers (Small-LR). In these two settings, all three systems succeed, with X-MoE delivering 27.33 TFLOPS (versus 27.08 and 28.26 for DeepSpeed-MoE and Tutel at Small-SR, and 62.51 TFLOPS versus 52.15 and 64.00 at Small-LR). These results confirm that X-MoE’s memory-efficient designs, especially our PFT-based expert routing, enable larger configurations under tight GPU memory constraints, with only a modest throughput trade-off due to the extra padding-free GEMM transforms required for maximal memory reuse on NVIDIA hardware.

\subsection{Implementation Validation}
\label{subsec:eval-impl-val}
\begin{figure}[!ht]
    \centering
    \includegraphics[width=0.75\linewidth]{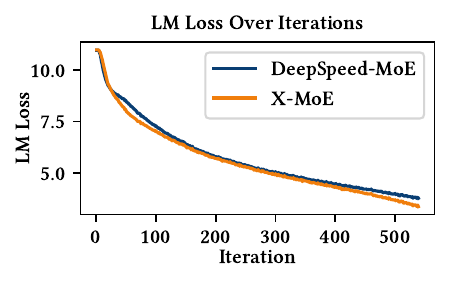}
    \caption{Loss validation with DeepSpeed-MoE and \name.}
    \label{fig:convergence-main}
    \Description{}
\end{figure}

To verify the correctness of \name, we compare its training loss curve against DeepSpeed-MoE on the 10.1B MoE model. The experiment is conducted on 16 GPUs with EP=8 and ZeRO DP enabled. In this setting, we confirm that \name closely tracks the convergence behavior of DeepSpeed-MoE, a production grade-implementation, as shown in \fref{fig:convergence-main}. 
This confirms that \name provides numerical convergence while enabling new system optimizations for scaling MoEs. We also investigate why the two curves do not match exactly and find that it is caused by a subtle difference in token-dropping logic. In DeepSpeed-MoE, a token is dropped from an expert if its routing score is negative, regardless of whether the expert's capacity has been exceeded. In contrast, \name only drops tokens when they exceed expert capacity. As a result, \name retains more tokens per batch, which might lead to its slightly lower loss under the same token consumption budget.   

\section{Related Work}

\noindent
\textbf{\emph{MoE Inference Frameworks.}} SGLang, vLLM, and TensorRT-LLM \cite{sglang, vllm, tensor-rt, sida-moe} are general inference frameworks that can also serve MoEs. SGLang provides optimized gather and scatter kernels in triton that are hardware agnostic; however, these kernels leverage block-sparse primitives and incur padding. vLLM provides optimized hardware-agnostic triton kernels via its FlashInfer \cite{flash-infer} backend. However, currently only MoEs which activate up to 7B parameters are supported. TensorRT-LLM is NVIDIA's LLM inference engine, but it is tightly coupled to the NVIDIA ecosystem. Moreover, none of these frameworks solve the activation memory explosion of large $A_{dispatch}$ and $A_{combine}$ tensors during MoE training.

\noindent
\textbf{\emph{Efficient Communication Primitives.}} DeepEP~\cite{deepep2025} is an open-source efficient EP implementation by DeepSeek, relying on intrinsics available only on NVIDIA Hopper GPUs. TCCL~\cite{tccl} modifies NVIDIA's NCCL to specifically optimize ring-based collectives on systems where the predominant interconnect is PCIe. Both techniques are tightly coupled to the NVIDIA ecosystem. Centauri~\cite{centauri} introduces an automated way to uncover good schedules where computation is overlapped with communication in heterogeneous environments by decomposing a training task hierarchically into multiple tiers. Unlike these works, \name focuses on system-level optimizations that enable scalable training of expert-specialized MoEs on non-NVIDIA platforms.

\label{sec:7_related}

\section{Conclusion}
\label{sec:conclusion}

In this paper, we have taken a leap forward in designing an MoE training system \name to scale expert-specialized MoEs, an increasingly popular model class. With techniques like padding-free MoE training pipeline with cross-platform kernels, redundancy-bypassing dispatching, and hybrid parallelism with sequence shard\-ed MoE blocks, \name enables training of massive MoEs on AMD-based HPC platforms while achieving high throughput, offering a system blueprint to train emerging expert-specialized MoEs on today's HPC platforms.

\section*{Acknowledgements}

We sincerely appreciate the insightful feedback from the anonymous reviewers. We also thank Emily Herron, Junqi Yin, and Hao Lu from ORNL for their useful discussion of this research. This research was supported by the National Science Foundation (NSF) under Grant No. 2441601. This manuscript has been authored by UT-Battelle, LLC under Contract No. DE-AC05-00OR22725 with the U.S. Department of Energy. The United States Government retains and the publisher, by accepting the article for publication, acknowledges that the United States Government retains a non-exclusive, paid-up, irrevocable, world-wide license to publish or reproduce the published form of this manuscript, or allow others to do so, for United States Government purposes. The Department of Energy will provide public access to these results of federally sponsored research in accordance with the DOE Public Access Plan (http://energy.gov/downloads/doe-public-access-plan). This research used resources at the Oak Ridge Leadership Computing Facility which is a DOE Office of Science User Facility. The work also utilized the Delta and DeltaAI system at the National Center for Supercomputing Applications (NCSA) through allocation CIS240055 from the Advanced Cyberinfrastructure Coordination Ecosystem: Services \& Support (ACCESS) program, which is supported by National Science Foundation grants \#2138259, \#2138286, \#2138307, \#2137603, and \#2138296. The Delta advanced computing resource is a collaborative effort between the University of Illinois Urbana-Champaign and NCSA, supported by the NSF (award OAC 2005572) and the State of Illinois. UIUC SSAIL Lab is supported by research funding and gift from Google, IBM, and AMD.



\bibliographystyle{ACM-Reference-Format}

\clearpage
\appendix
\section{Evaluation Setup Details}
\label{sec:detail-setup}

Our evaluation is based on DeepSpeed version 0.15.5 and DeepSpeed-Megatron. We use PyTorch version 2.2.0 and AMD ROCm version 5.7.1. The peak device throughput of two MI250X GCDs is 383 TFLOPS, and the per-effective-GPU peak throughput is 191.5 TFLOPs. For cross-node communication on Frontier, we use the AWS-OFI-RCCL plugin \cite{Kheria2024ROCm} to enhance inter-node connectivity, which maps RCCLs connection-oriented transport APIs to libfabric’s interface. We use libfrabric version 1.20.1. We set environmental variables \texttt{CUDA\_DEVICE\_MAX\_CONNECTIONS=1} and \texttt{NCCL\allowbreak\_NET\allowbreak\_GDR\allowbreak\_LEVEL=3}
 for better RCCL efficiency, as recommended by \cite{frontier}. For DeepSpeed-MoE and DeepSpeed-TED, we use the DeepSpeed library version 0.15.5. 
Since Tutel is not provided as the end-to-end training pipeline, we integrate its MoE layer implementation from Tutel library version 0.3 into the DeepSpeed library. In end-to-end experiments, we refer to Tutel for this integration. 

\section{Implementation of PFT Training Pipeline}
\subsection{Conventional MoE Pipelines with Zero-Padding}
\label{sec:zero-padded-pipeline} 

\begin{figure}[!ht]
    \centering
    \includegraphics[width=1\linewidth]{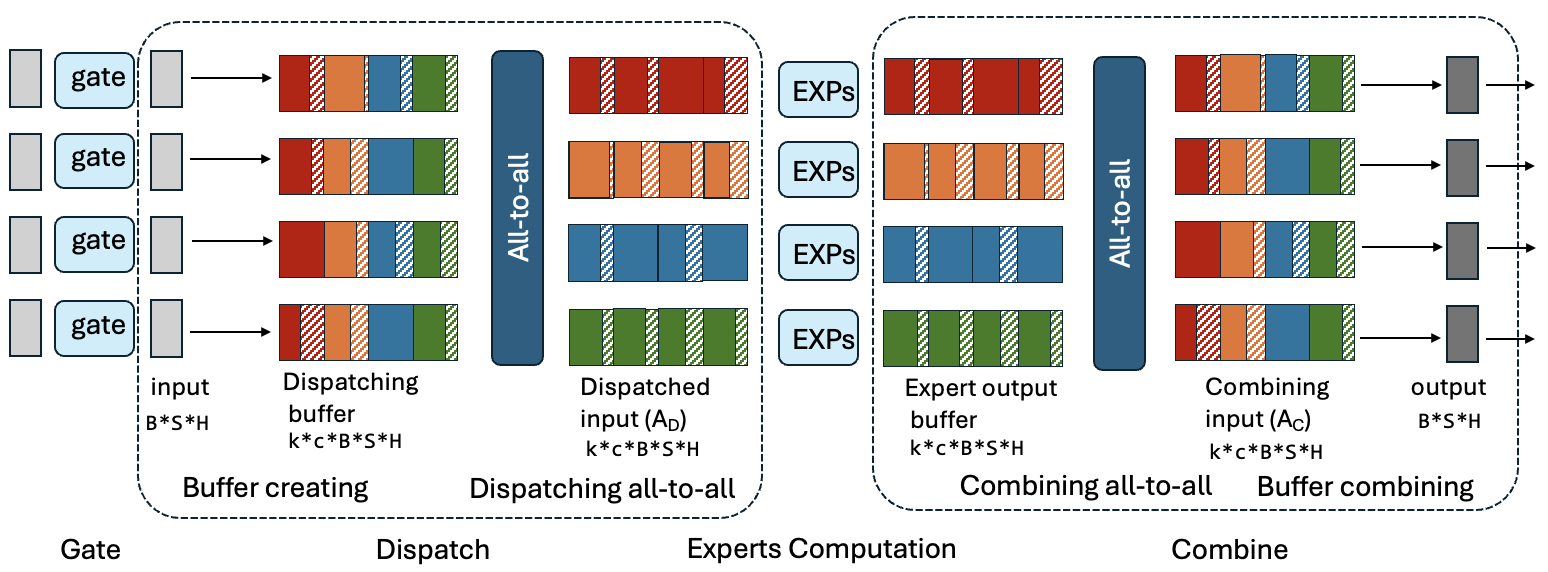}
    \caption{DeepSpeed-MoE training pipeline.}
    \label{fig:ds-moe-training-pipeline} 
    \Description{}
\end{figure}

Conventional MoE training frameworks implement each stage of the MoE pipeline, as depicted in \fref{fig:ds-moe-training-pipeline}, via fast batched matrix multiplication (matmul) primitives. However, these primitives place a constraint: requiring the same number of tokens routed to each expert, which does not hold during training. To handle the dynamic assignment of tokens to experts, these frameworks introduce an \textit{expert-capcity} factor, $C$. 
When token counts fall short of $C$, expert buffers are zero-padded, and when they exceed $C$, they are dropped, resulting in equal sized expert input-buffers. 

\emph{Gating.} During the gating stage, a dispatching mask, \texttt{dispatch\_mask} of size \texttt{[S, E, C]} is constructed. The entry \texttt{dispatch\_mask[t, e, c]} is either 1 or 0 indicating if the \texttt{t$^{\text{th}}$} token is routed to the \texttt{c$^{\text{th}}$} position in expert \texttt{e}'s buffer. A token-dropping mask is applied over \texttt{dispatch\_mask} to additionally drop tokens. 

\emph{Dispatch, MLP and Combine.} Each of the dispatch, MLP and combine stages leverage matmuls on input token-buffers to process tokens. First, during the dispatch stage, each worker uses an \texttt{einsum} operation on the dispatching mask and input tokens to correctly place each token into its respective experts buffers. The expert buffers are \texttt{[E, C, H]}-sized. If less than \texttt{C} tokens are placed in an expert's buffer, the unused slots are zero-padded. Figure ~\ref{fig:gshard} illustrates the dispatch process and how zero-padding is introduced into an expert's input-buffer at this stage. Next, an even all-to-all communication exchanges token buffers across devices, correctly routing each token to the devices its experts reside on. Each expert then operates on its input token-buffers in parallel. Finally, another all-to-all re-exchanges the tokens to their original device to generate the final output of the layer. Importantly, the zero-padding introduced in the dispatch stage is retained across each all-to-all and expert compute. This increases both the communication volume and activation memory of MoE training.

\subsection{PFT Construction}
\label{sec:token-dropping-pft}
The PFT construction routine proceeds in two stages. In the first stage, we flatten and sort the incoming \texttt{top\_experts} array (lines 11-12), which contains the token to expert assignments generated by the gating function. In the second stage, we determine which tokens are dropped (lines 15-26); using this information, we construct the \metadata by pruning out the dropped tokens from the unfiltered \texttt{token\_ids} and \texttt{expert\_ids} (lines 25-26).
Our token dropping strategy is parameterized by a variable: \texttt{max\_token\_count}, indicating the maximum number of tokens a worker can route to an expert. This parameter is decided prior to training by the user. We use this to select the top \texttt{max\_token\_count} tokens per expert in \texttt{gate$_{\text{out}}$} (the \texttt{[S, H]}-sized token-buffer output of the gating kernel), ranked by their respective \texttt{combine\_weights}. This retains \textit{at-most} \texttt{max\_token\_count * E} tokens. We achieve this by first flattening and sorting the \texttt{top\_experts} array according to their respective \texttt{combine\_weights} (Listing 1 lines 15-17). Then, we one-hot-encode the last dimension, forming \texttt{one\_hot\_enc} (line 19). \texttt{one\_hot\_enc[i, j]} is either 1 or 0, indicating whether token \texttt{i} is routed to expert \texttt{j} or not respectively. Next, we cumulatively sum across the inner dimension, producing \texttt{rank\_in\_expert} (line 20). Here, \texttt{rank\_in\_expert[i, j] = b} indicates that \texttt{b} tokens in \texttt{0...i} have been routed to expert \texttt{j}. From this array we determine if token \texttt{i} is routed to expert \texttt{j} if and only if \texttt{rank\_in\_expert[i, j] < max\_token\_count} and \texttt{one\_hot\_encoded[i, j] = 1}, producing the \texttt{retained\_token\_ids} array (lines 21-24). Finally, using this information we filter out the dropped tokens in the \texttt{token\_ids} and \texttt{expert\_ids} arrays (lines 25-26). 

We optimize our token-dropping strategy by observing that applying a \texttt{cumsum} on the inner dimension of a tensor is slow as memory requests are not coalesced. To fix this, we manually create the \texttt{one\_hot\_enc} tensor to be \texttt{[E, S*K]} sized and instead apply the \texttt{cumsum} on the outer dimension. By transposing the data-layout, this optimization accelerates the combined sum of the gating and PFT construction process by 10x. 
After token dropping, we compute the length of each expert's segment in the dispatched buffer with a \texttt{histogram} producing the \texttt{tokens\_per\_expert} \metadatum (line 28). Finally, we return all the generated \metadata (line 30), concluding the PFT construction process.

\subsection{Padding-free MoE-layer}
\label{sec:padding-free-pipeline}
In this section, we give an in-depth description of how the MoE layer is modified to incorporate the PFT structure and remove zero padding. 

\emph{Padding-free Gating, Dispatch, MLP and Combine.} Although zero-padding is not introduced during gating, introducing the PFT enables us to eliminate the large auxiliary data-structures produced in conventional MoE frameworks during this stage, saving memory. Our modified gating stage, similar to conventional MoE frameworks, first projects the input tokens of shape \texttt{[S, H]} to logits of shape \texttt{[S, E]} and selects the top-$k$ experts per token, resulting in two outputs: \texttt{top\_experts} and \texttt{combine\_weights}, both of shape \texttt{[S, K]}. However, unlike conventional MoE frameworks, no \texttt{dispatch\_mask} is produced. Instead the \texttt{top\_experts} array is consumed by the PFT construction routine (line 11) 
to produce the necessary \metadata required for the dispatch, MLP and combine stages.

\emph{Dispatch.} Next, the \texttt{token\_ids} and \texttt{expert\_ids} arrays (output form the PFT construction routine) alongside a triton-gather kernel will create the dispatch matrix. This proceeds in two stages. First, we instantiate an empty dynamic-sized buffer, \texttt{dispatch$_{\text{in}}$} (the dispatch matrix) which is of shape \texttt{[B, H]}, where $B$ is the length of the \texttt{token\_ids} array and $H$ is the model hidden dimension. Next, \texttt{dispatch$_{\text{in}}$}, alongside the \texttt{expert\_ids} and \texttt{token\_ids} arrays will be input into a custom triton-gather kernel (more details in \sref{subsec:pipeline-backend}) which will copy the tokens from the output of the gating-kernel, \texttt{gate$_{\text{out}}$} to the correct indexes in \texttt{dispatch$_{\text{in}}$}, according to the indexing specified by the \texttt{token\_ids} array. Figure ~\ref{fig:pft} shows how the PFT and its \metadata create the \texttt{dispatch\_in} matrix. Following the creation of the dispatch matrix, we then exchange tokens between devices via an uneven \texttt{alltoall} with no zero-padding communicated, reducing total communication volume. Again, this proceeds in two stages. First, we exchange the \texttt{tokens\_per\_expert} \metadatum, allowing each device to compute the number of \textit{inbound} tokens routed to it, which we denote as $B_{exp}$. We use this information to create a new buffer, \texttt{dispatch$_{\text{out}}$}, of size \texttt{[B$_{\text{exp}}$, H]}. Next, we exchange the tokens using the \texttt{alltoallv} backend, populating the \texttt{dispatch$_{\text{out}}$} buffer. At the end of this stage the PFT's token-buffer, \texttt{x}, is assigned to \texttt{dispatch$_{\text{out}}$} with its \texttt{tokens\_per\_expert} \metadatum updated according to the contents of \texttt{dispatch$_{\text{out}}$}. The rest of the \metadata are unmodified.

\emph{MLP.} Next, the modified PFT produced by the dispatch stage, containing the necessary tokens that experts residing on a device will process, is consumed by the MLP layer. The MLP layer implements two padding-free GeMMs that consume the PFT token buffer, \texttt{x} (previously assigned to the output of the dispatch stage, \texttt{dispatch$_{\text{out}}$}), as well as the \texttt{tokens\_per\_expert} metadata, launching a sequential GeMM to compute each MLP layer (described in \sref{subsec:pipeline-backend}). The output of the second GeMM is of size \texttt{[B$_{\text{exp}}$, H]}, denoted as \texttt{mlp$_{\text{out}}$}. At the end of this stage the PFT' token-buffer, \texttt{x}, is assigned to \texttt{mlp$_{\text{out}}$} with the rest of \metadata unmodified. 

\emph{Combine.} Finally, the modified PFT produced by the MLP stage is input to the combine stage, which communicates the tokens in the PFT's token-buffer, \texttt{x}, back to their original respective device. In this stage, we first use an \texttt{alltoall} to exchange tokens to the correct device, re-forming a \texttt{[B, H]}-sized matrix, denoted as \texttt{combine$_{\text{in}}$}. Then, a custom scatter-kernel (described in \sref{subsec:pipeline-backend}) will consume the original \texttt{token\_ids}, \texttt{expert\_ids} and \texttt{combine\_weights} \metadata and reorder the tokens in \texttt{combine$_{\text{in}}$} according to the \texttt{token\_ids} array (undoing the effect of the gather-kernel) while multiplying each token by its respective value in \texttt{combine\_weights}. This creates a new \texttt{[S, H]}-sized buffer (\texttt{S} is the original token-count input to the MoE layer) as the final output, concluding the combine stage.

\subsection{Gather, Scatter \& Sequential GeMM} 
\label{sec:scatter-kernel}

\emph{Gather and Scatter Kernel.} The gather kernel, written in triton, is used to copy and reorder tokens from the output of the gating kernel, \texttt{gate$_{\text{out}}$} (\texttt{[S, H]}-sized), to the dispatch buffer, \texttt{dispatch$_{\text{in}}$} (\texttt{[B, H]}-sized), according to the indexing specified within the \texttt{token\_ids} (\texttt{[B]}-sized) tensor. It performs the operation \newline \texttt{dispatch$_{\text{in}}$[i, :] = gate$_{\text{out}}$[token\_ids[i], :]}, implementing a classical gather kernel. The kernel launches \texttt{B} thread-blocks, each containing 256 threads, with thread-block \texttt{bi} responsible for copying \texttt{gate$_{\text{out}}$[token\_ids[bi], :]} to \texttt{dispatch$_{\text{in}}$[bi, :]}. Each thread-block loops over the hidden dimension, \texttt{H/256} times, with consecutive threads assigned to move consecutive values in \texttt{gate$_{\text{out}}$[token\_ids[bid]]} to consecutive memory locations in \texttt{x[token\_ids[bid]]}. This ensures that memory requests are coalesced despite the irregular memory access patterns that arise due to nested tensor indexing in the expression \texttt{gate$_{\text{out}}$[token\_ids[i], :]}. 
The scatter kernel reverses this operation, sending tokens back to their original positions in the sequence and applying the corresponding routing weights:
\texttt{combine$_{\text{in}}$[token\_ids[i], :] = mlp$_{\text{out}}$[i, :] $\times$ \texttt{combine$\_{\text{weights}}$}}, implementing a classical scatter kernel. 
Unlike the gather kernel, the scatter kernel's irregular memory access patterns arises due to \textit{writing} to the output buffer, \texttt{combine$_{\text{in}}$}. However, similar to the gather kernel, memory coalescing is ensured by scheduling threads to operate on consecutive memory locations across the hidden-dimension.
Unlike prior work like Megablocks \cite{gale2023megablocks} that use scatter and gather kernels with zero-padded data, our gather and scatter kernels operate on padding-free data.

\emph{Sequential GeMM based expert computation.} The sequential GeMM implements the two-layers of the MLP without the need for any padding. It consumes the dispatch matrix, \texttt{dispatch$_{\text{out}}$} and the \texttt{tokens\_per\_expert} \metadatum. It uses the \texttt{tokens\_per\_expert} \metadatum to correctly track which tokens in the \texttt{dispatch$_{\text{out}}$} buffer should be multiplied by which expert. 
On each device, we launch a sequence of $E_{\text{local}}$ GeMMs (equal to the number of experts assigned to the device), each compute one expert's tokens. 
The \texttt{$\text{i}^{\text{th}}$} expert processes tokens:
\texttt{dispatch$_{\text{out}}$[sum(tpi[:i+1]):sum(tpi[:i+2])]}, where \texttt{tpi} is the \texttt{tokens\_per\_expert} \metadatum.

\subsection{Complexity Analysis}
We compare the memory and computational costs of our proposed PFT dispatching strategy against the standard GShard‐style approach.  Let $b$ denote the batch size, $s$ the sequence length, $h$ the hidden dimension, $k$ the number of experts per token, and $c$ the GShard capacity factor.  By maintaining only a token‐level buffer of size $k$ per input, PFT requires $O(kbsh)$ memory, since no zero padding to a fixed capacity is performed.  In contrast, GShard‐style dispatching must pad each buffer to the worst‐case capacity and allocate intermediate position‐encoding matrices, yielding $O(ckbsh) + O(ckb^{2}s^{2})$ memory overhead.  On the computation side, PFT achieves $O(kbsh)$ by directly processing only the nonzero entries, whereas GShard’s use of large, padded tensors incurs $O(ckb^{2}s^{2}h)$ due to costly matrix multiplications with zero‐padded buffers.  These complexity bounds explain PFT’s superior memory footprint and compute efficiency in large‐scale MoE deployments.

\section{Analysis of Hybrid Parallelism Strategy on Frontier}
\subsection{EP/DP Placement Strategy: EP-First vs. DP-First?}
\label{sec:ep-vs-dp}

One key decision in training MoEs on large GPU clusters lies in how we combine EP and DP across devices. While both EP and DP are needed to scale model size and training throughput, they present conflicting locality goals when placed across the same set of GPUs. 
We refer to this tension as the Locality-aware EP vs. Replica-aware DP tradeoff: (1) \emph{Locality-aware EP} places as many different experts closely (e.g., within a node) to eliminate expensive inter-node \texttt{alltoall} communication, which minimizes EP token routing cost. (2) \emph{Replica-aware DP} replicates the same experts on GPUs within the same node to reduce inter-node communication for gradient synchronization.

These two goals are mutually exclusive: maximizing expert diversity per node for EP inherently increases the number of distinct parameters, and thus DP communication cost. In contrast, grouping the same expert replicas for DP forces token routing for EP to go across nodes. This leads to two configurations: \emph{EP-First} placement (EP-then-DP) and \emph{DP-First} placement (DP-then-EP). 
Intuitively, grouping experts closely and using DP to scale the MoE training as EP-then-DP may help reduce inter-node latency from expensive \texttt{alltoall} calls. Indeed, this is the strategy existing MoE training systems such as DeepSpeed-MoE use for scaling MoEs. However, the optimal placement depends on both the model and the hardware topology. For small MoEs, locality-aware EP may win, because EP \texttt{alltoall} dominates the communication cost. For relatively large MoEs, replica-aware DP actually becomes more appealing, because DP needs to synchronize data volume linear with respect to the number of parameters. 

As a concrete example, consider training an MoE on 64 GPUs (8 nodes $\times$ 8 GPUs per node). Suppose the model has 8 experts, and we distribute them with EP=8. 
The EP-First strategy places all 8 experts within each node, and replicate this expert set across all 8 nodes. DP-First spreads the 8 experts across 8 nodes, placing one unique expert per node and replicating each expert across the 8 GPUs within the node. In EP-First, each node holds a full replica of the expert set. During DP gradient synchronization, each parameter appears once per node, which requires inter-node communication to average gradients across all replicas. For large MoEs, this results in high inter-node bandwidth pressure, which is expensive on HPC clusters such as Frontier, which only has 25GB/s inter-node bandwidth. In contrast, DP-First co-locates all replicas of the same expert within a node. Therefore, most DP communication happens within a node, where Frontier offers up to 200GB/s bandwidth. By reordering the placement, favoring DP-First, we shift DP communication from slow inter-node links to fast intra-node connects. This change leads to significant performance gains on HPC platforms with hierarchical bandwidth asymmetry.

\subsection{Tradeoff Analysis Between \ssmb and TED}
\label{sec:ted-vs-ssmb}

We analyze the overhead and gain of using \ssmb compared with tensor-expert-data (TED) parallelism. In the following analysis, we consider two cases:
 \begin{itemize}
     \item 
     Applying \ssmb with TP=$G$.
     \item 
     Apply tensor-expert-data parallelism (TED) with TP=$G$.
\end{itemize}
 
For the two cases, we assume an identical EP size and DP size.

In \ssmb, we distribute the all-to-all communication volume across the AS group, and the total communication volume in the EP group remains the same.
To reconstruct the tensors, we need an all-gather at the end of the MoE layer. Also, in the backward pass, we need another all-gather to get the full gradient of the input tensor. Thus, the extra communication volume is $O(BSH)$ on each device, which is at the same magnitude of TED.

For memory, the \ssmb reduces the activation memory by $G$ times. Assuming using half precision training, the saved activation memory per-device is
\begin{equation}
A_{\text{saving}}=4ckSH\frac{G - 1}{G}  
\label{eq:mem-save}
\end{equation}
bytes, where $S$ is sequence length and $c$ is the capacity factor.
Compared with TED, it does not distribute the model parameter and increases the model memory. Assume that we apply ZeRO-1 DP to scale the model, and the optimizer states of each expert are distributed to devices in the same DP group.  
 The model states memory increasing by applying \ssmb instead of TED is 
$$M_{\text{cost}}=\frac{E}{\text{EP size}}\cdot 8H_{FFN}H \frac{G-1}{G   }$$ bytes.
Since we can choose EP size freely up to the number of experts $E$, the lowest cost bound is 
\begin{equation}
M_{\text{mincost}}=8H_{FFN}H \frac{G-1}{G} 
\label{eq:mem-cost}
\end{equation}

\begin{figure}[H]
    \centering
    \includegraphics[width=1\linewidth]{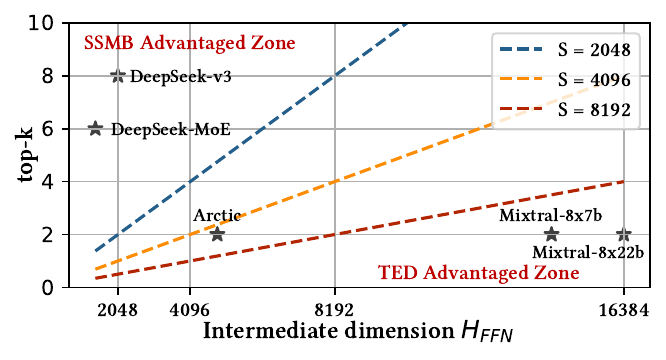}
    \caption{The memory saving advantage regions of \ssmb and TED.}
    \label{fig:tsp-vs-ted}
    \Description{}
\end{figure}

For more intuitive understanding, we calculate the ratio $r$ of the configurations from multiple popular MoE models, including Mixtral-8x7b, Mixtral-8x22b, DeepSeek-MoE, DeepSeek-v3, and Arctic, and plot in \fref{fig:tsp-vs-ted}. The Mixtral series models are conventional MoE models, and the DeepSeek series models are typical emerging-style MoE models. The Arctic is mixed - it uses the fine-grained experts without a large top-$k$. In this figure, the upper region is the advantage region of \ssmb, and the lower region is the advantage region of TED. The region borderline depends on the sequence length $S$ and capacity factor $c$ choice. For simplicity, we choose capacity factor $c=1$, and plot the advantage region borderline for three sequence lengths, $S=2048, 4096, 8192$ as examples. The figures show that for all three $S$ choices, DeepSeek series models save more memory from \ssmb than TED. Also, for all three $S$ choices, Mixtral series models save more memory from TED than \ssmb. As for Arctic, the best strategy choice depends on the training sequence length. The longer the sequence length, the better to choose \ssmb rather than TED.

\section{Training Sensitivity at Scale}
\label{sec:training-sensitivity}

We observe that on large supercomputers, network performance becomes increasingly sensitive to system scale. As we increase the number of GPUs used for training, the system behavior begins to be influenced by dynamics beyond just memory usage and communication volume.  To quantify this effect, we profile the all-to-all communication latency on Frontier while scaling from 8 to 1024 GPUs. The results reveal three distinct regions: (i) latency increases from 8 to  GPUs, (ii) latency remains relatively stable from 32 to 256 GPUs, (3) it rises sharply beyond 256 GPUs. We hypothesize that this sharp latency increase is due to Frontier's hardware topology: a single rack contains up to 256 GPUs, while communication beyond this threshold generates cross-rack traffics, which is more prone to network congestion from other concurrently running jobs. Based on the profiling result, we limit the EP size of our training strategies up to $256$.

\begin{figure}[!ht]
    \centering
    \includegraphics[width=\linewidth]{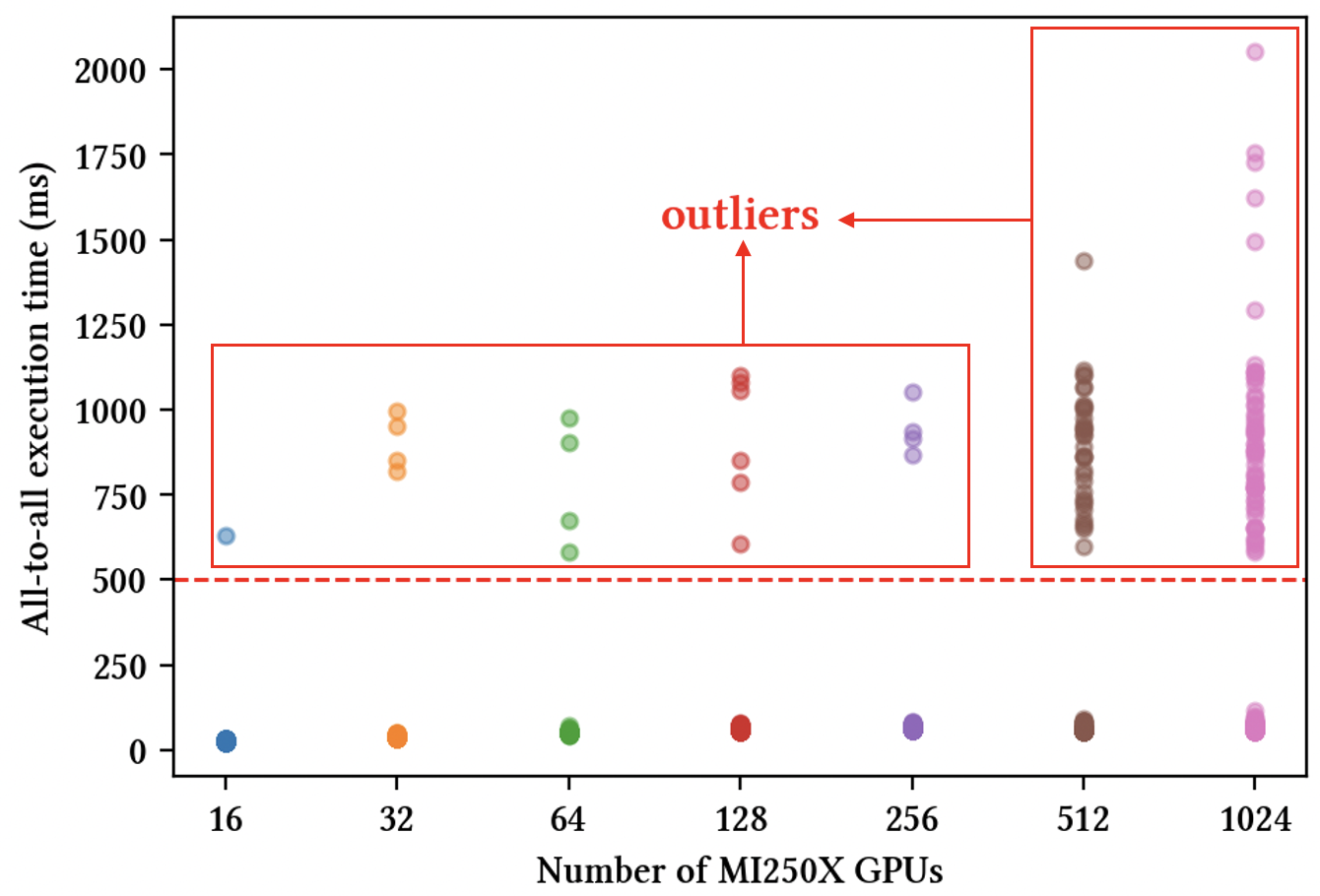}
    \caption{The all-to-all collective time characterization across 1000 runs. Most of the runtime aggregates at the bottom, $<100$ ms. Above them are the outliers, which occur frequently for 512 and 1024 GPUs.}
    \label{fig:outliers}
    \Description{}
\end{figure}

We show the the all-to-all collective time of 1000 runs in \fref{fig:outliers} by plotting each runtime as a scatter point on the figure. It reveals the increasing frequency of outlier all-to-alls, which have a per-collective time $>500$ ms for 512 and 1024 GPUs. 




\begin{figure}[!ht]
    \centering
    \includegraphics[width=0.8\linewidth]{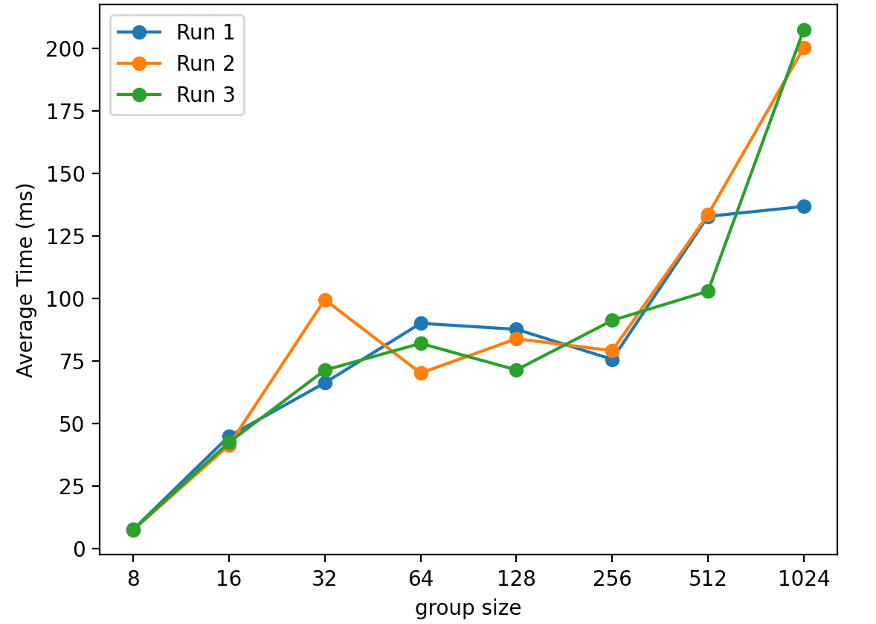}
    \caption{The average all-to-all time of the MoE training workload across varying GPU scale.}
    \label{fig:enter-label}
    \Description{}
    
\end{figure}


\section{Additional Model Scaling Results}
\label{sec:extra-comparison}

\begin{figure*}[!ht]
    \centering
    \includegraphics[width=\linewidth]{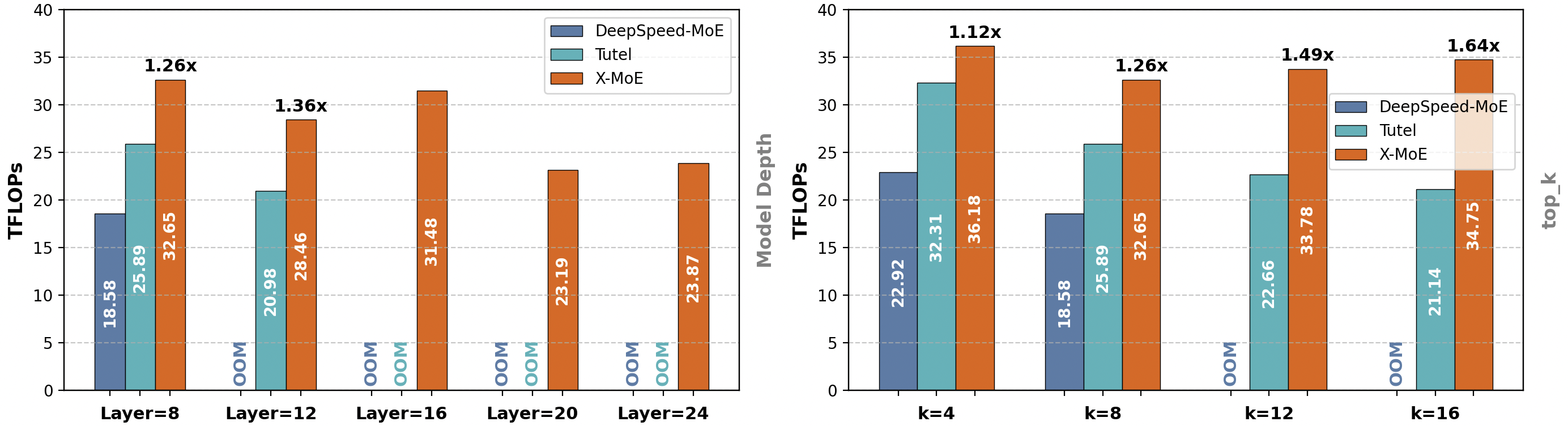}
    \caption{Scaling efficiency of \name on 256 GPUs with varying model configurations. (Left) Training throughput when increasing the number of MoE layers. \name consistently achieves $>$22TFLOPs throughput from 8 to 24 layers, while baselines run out of memory beyond 8 layers. (Right) Training throughput when increasing top-$k$ values in token routing. \name scales better with increasing $k$.}
    \label{fig:scale-layers-and-k}
    \Description{}
\end{figure*}


In this section, we present additional results that further demonstrate the scalability of \name across various model configurations on 256 GPUs. These experiments include:
\begin{itemize}
    \item Scaling total and activated parameter count by increasing the number of layers.
    \item Scaling activated parameter count by increasing the top-$k$ value in MoE routing.
\end{itemize}


We use the \emph{Large} model configuration as a base and vary: 
\begin{itemize}
    \item The number of layers in \{8, 12, 16, 20, 24\}.
    \item The top-$k$ value in \{4, 8, 12, 16\} with the number of layers fixed. 
\end{itemize}


In all configurations, we compare the training throughput of DeepSpeed-MoE, Tutel, and \name. For \name, we set EP=64 and vary TP between 1 and 2 depending on memory capacity.

\noindent
\emph{Scaling by model depth and top-$k$.} We show the performance comparison when scaling the model along the depth dimension in \fref{fig:scale-layers-and-k} (left).
The results show that \name can efficiently scale both the model depth and the top-$k$ better than baselines. In the experiment on model depth, both DeepSpeed-MoE and Tutel run out of memory as the number of layers exceeds 16, while \name consistently achieves over 23 TFLOPs training throughput from 8 to 24 layers. 
This indicates that the existing parallelism strategies do not perform well in scaling the emerging MoE models.
In contrast, \name effectively scales the number of layers with optimizations introduced in \sref{sec:design}. 

\name also scales better with increasing top-$k$ values. For smaller $k$ values (e.g., $k=4$), the throughput gains over Tutel are moderate (1.12$\times$), but as $k$ increases, \name achieves up to 1.64$\times$ higher throughput when $k=16$. 
As discussed in \sref{subsec:communication_analysis}, the all-to-all communication volume is linear to $k$, which becomes significant as $k$ increases 
In contrast, \name mitigates this overhead by reducing inter-node communication through padding-free pipeline and \rbd.
\label{sec:extra-correctness}

\end{document}